\documentclass[journal]{IEEEtran}
\usepackage{graphicx}
\usepackage{tikz}
\usepackage{tcolorbox}
\usepackage{pgfplots}
\pgfplotsset{compat=1.18}
\usetikzlibrary{positioning}
\usetikzlibrary{arrows.meta, positioning}
\usepackage[utf8]{inputenc}
\usepackage{booktabs}
\usetikzlibrary{mindmap, shadows}
\usepackage{lipsum}  
\usepackage[ruled,vlined]{algorithm2e}
\usepackage{amsmath,amssymb}
\usepackage{amsmath}
\usepackage{siunitx}
\usepackage{ragged2e}
\usepackage{siunitx}
\usepackage{tabularx}
\usepackage{adjustbox}
\usepackage{caption}
\usepackage{graphicx}
\usepackage{subcaption}
\usepackage{float}

\usepackage{amsmath,amssymb,mathtools} 

\usepackage{siunitx}
\usepackage{graphicx}
\usepackage{booktabs}

\usepackage{microtype}
\usepackage[dvipsnames]{xcolor}

\usepackage[hidelinks]{hyperref}
\usepackage[nameinlink,capitalise]{cleveref} 

\usepackage{algpseudocode}

\usepackage{colortbl}
\usepackage{xcolor}
\usepackage{framed}
\usepackage{booktabs}
\usepackage{pifont}
\usepackage{siunitx}
\usepackage{url}
\usepackage{tabularx}     
\usepackage{enumitem} 

\newcommand{\event}[1]{\mathsf{#1}}

\definecolor{rqgray}{gray}{0.90} 
\newcommand{\maxnum}[1]{\textbf{\underline{\num{#1}}}}

\usepackage{amsmath, amssymb}
\usepackage{xcolor}
\usepackage{booktabs}
\usepackage{caption}
\usepackage{pgfplots, pgfplotstable}
\pgfplotsset{compat=1.18}
\usepackage{sansmath}
\usetikzlibrary{arrows.meta}

\pgfplotsset{
  myaxis/.style={
    grid=both, grid style={line width=.1pt, draw=gray!25},
    major grid style={line width=.2pt, draw=gray!45},
    tick style={black},
    tick label style={/pgf/number format/fixed, font=\small},
    label style={font=\small},
    legend style={font=\small, cells={anchor=west}},
  }
}

\newcommand{\TopN}{20} 
\pgfplotsset{
  keeptoprows/.style={
    row predicate/.code={%
      \ifnum##1<\TopN\relax
      \else
        \pgfplotstableuserowfalse
      \fi
    }
  }
}

\usepackage{array}
\newcolumntype{L}[1]{>{\raggedright\arraybackslash}p{#1}} 
\newcolumntype{Y}{>{\raggedright\arraybackslash}X}         
\definecolor{grayrow}{gray}{0.95}
\renewcommand{\arraystretch}{1.1} 
\usepackage{amssymb}    
\usepackage{pifont}     
\usepackage{xcolor}     

\definecolor{cyanblue}{RGB}{224,238,255}

\ifCLASSINFOpdf
\else
\fi
\begin{document}
%
\title{\textit{AgentDrive}: An Open Benchmark Dataset for Agentic AI Reasoning with LLM-Generated Scenarios in Autonomous Systems}

\author{
Mohamed~Amine~Ferrag$^{*\S}$,~\IEEEmembership{Senior~Member,~IEEE}, 
Abderrahmane~Lakas$^{*}$,~\IEEEmembership{Senior~Member,~IEEE},
and~Merouane~Debbah$^{1}$,~\IEEEmembership{Fellow,~IEEE}%
\thanks{$^{*}$Department of Computer and Network Engineering, College of Information Technology, United Arab Emirates University, Al Ain, United Arab Emirates.}%
\thanks{$^{1}$Khalifa University of Science and Technology, Abu Dhabi, United Arab Emirates.}%
\thanks{$^{\S}$Corresponding author: \texttt{mohamed.ferrag@uaeu.ac.ae}}%
}

%
%

\markboth{ }%
{Shell \MakeLowercase{\textit{et al.}}: Bare Demo of IEEEtran.cls for IEEE Journals}
%



\maketitle


\begin{abstract}
The rapid advancement of Large Language Models (LLMs) has sparked growing interest in their integration within autonomous systems to enable reasoning-driven perception, planning, and decision-making. However, evaluating and training such agentic AI models remains challenging due to the lack of large-scale, structured, and safety-critical benchmarks. This paper introduces AgentDrive, an open benchmark dataset containing 300,000 LLM-generated driving scenarios designed for the training, fine-tuning, and evaluation of autonomous agents under diverse conditions. AgentDrive formalizes a factorized scenario space across seven orthogonal axes—scenario type, driver behavior, environment, road layout, objective, difficulty, and traffic density—and employs an LLM-driven prompt-to-JSON pipeline to produce semantically rich, simulation-ready specifications that are validated against physical and schema constraints. Each scenario undergoes simulation rollouts, surrogate safety metric computation, and rule-based outcome labeling. To complement simulation-based evaluation, we introduce AgentDrive-MCQ, a 100,000-question reasoning benchmark spanning five reasoning dimensions—physics, policy, hybrid, scenario, and comparative—to systematically assess the cognitive and ethical reasoning of LLM-based agents. We conducted a large-scale evaluation of fifty leading LLMs on the AgentDrive-MCQ benchmark to measure their reasoning capabilities across these five dimensions, covering models such as GPT-5, ChatGPT 4o, Gemini 2.5 Flash, DeepSeek V3, Qwen3 235B, ERNIE 4.5 300B, Grok 4, Mistral Medium 3.1, and Phi 4 Reasoning Plus. Results reveal that while proprietary frontier models dominate in contextual and policy reasoning, advanced open models are rapidly closing the gap in structured and physics-grounded reasoning. To support open science and reproducibility, we release the \texttt{AgentDrive} dataset (including labeled data), the AgentDrive-MCQ benchmark, evaluation scripts, and all related materials on GitHub: \url{https://github.com/maferrag/AgentDrive}.
\end{abstract}


%
\IEEEpeerreviewmaketitle

\section{Introduction}

The rise of large language models (LLMs) has sparked broad interest in applying them to autonomous driving (AD) due to their strong reasoning and conversational abilities \cite{yang2023llm4drive}. Researchers have introduced the concept of LLM4AD, designing AD systems that leverage LLMs for tasks ranging from perception and scene understanding to decision-making \cite{cui2024large,cui2024survey,gao2025foundation}. For example, Cui \textit{et al.} \cite{cui2024large} proposed an LLM4AD framework with a comprehensive simulation benchmark to evaluate how well LLMs follow driving instructions. Initial experiments in both simulation and real-world driving showed that LLMs can indeed enhance an autonomous vehicle’s understanding of complex environments and improve human–vehicle interactions. These findings, echoed by recent surveys \cite{yang2023llm4drive,cui2024survey}, suggest that combining LLMs with vision models could enable more open-world perception, logical reasoning, and adaptive learning than traditional rule-based or end-to-end systems. At the same time, the LLM4AD concept paper and surveys emphasize that this nascent field faces significant challenges, such as ensuring real-time performance and safety, which must be addressed as development continues.

Integrating LLMs into an autonomous driving agent raises new complexities in how the vehicle interprets instructions, interacts with humans, and adheres to traffic rules. One approach to this is DriVLMe, an LLM-based driving agent augmented with both embodied experiences (learned via a driving simulator) and social experiences from real human dialogues \cite{huang2024drivlme}. The goal of DriVLMe is to facilitate natural communication between humans and self-driving cars and to handle long-horizon navigation tasks through free-form dialogue. DriVLMe demonstrated competitive performance in open-loop simulations and closed-loop user studies; however, it also revealed several limitations, including unacceptably high inference times, imbalanced training data, and difficulty handling sudden environmental changes. To ensure LLM-driven vehicles make decisions that are not only effective but also legally and ethically aligned, other works focus on injecting domain knowledge and safeguards into the decision process. For example, Cai \textit{et al.} \cite{cai2024driving} present a retrieval-augmented reasoning framework where a Traffic Regulation Retrieval module automatically fetches relevant traffic laws and safety guidelines based on the vehicle’s situation, and an LLM then interprets these rules to assess each action’s legality and safety. This approach yields an interpretable decision-making pipeline that adapts to regional regulations and provides transparency into why the AI chooses a certain action. Similarly, Kong \textit{et al.} \cite{kong2024superalignment} introduce a multi-agent superalignment framework to enforce data security and policy compliance in LLM-driven cars. Their system safeguards sensitive vehicle data (e.g., precise locations, camera feeds) from potential leaks and filters the LLM’s queries and outputs, ensuring that driving commands do not violate safety rules or human values. These efforts demonstrate how researchers are enhancing LLM-based driving agents with embodied learning and alignment mechanisms, enabling autonomous vehicles to interact naturally while adhering to legal, safety, and trust constraints.

Beyond high-level dialogue and compliance, LLMs and multimodal models are being used to tackle core perception and planning challenges in autonomous driving. Bai \textit{et al.} \cite{bai20243d} argue that prior vision–language planning approaches, which tokenize only 2D images, cannot reliably perceive a 3D world. They propose a planner that uses DETR-style 3D perceptrons as tokenizers, feeding the LLM with 3D object queries from multi-view camera images. This strategy provided rich geometric context, yielding superior performance in 3D object detection and end-to-end planning on the nuScenes dataset, suggesting that 3D-tokenized LLMs could be key to more reliable path planning. Sah \textit{et al.} \cite{sah2025advancing} integrate deep learning models for traffic sign recognition and lane detection with a lightweight multimodal LLM that can reason about the scene. This hybrid system achieved near-perfect accuracy in clear conditions and maintained robustness under adverse weather and occlusions, demonstrating the LLM’s ability to interpret contextual clues. Other studies leverage LLMs for specialized tasks such as parking in mixed traffic \cite{jin2024large} and interpretable end-to-end control. Chen \textit{et al.} \cite{chen2024driving} fuse object-level vector modalities with LLM reasoning to provide explainable driving decisions, while Xu \textit{et al.} \cite{xu2024drivegpt4} propose DriveGPT4, which generates both low-level control signals and natural language explanations for each action. These advances highlight LLMs as powerful engines for enhancing perception, planning, and transparency in AD systems.

\begin{figure}[htbp]
    \centering
    \includegraphics[width=0.5\textwidth]{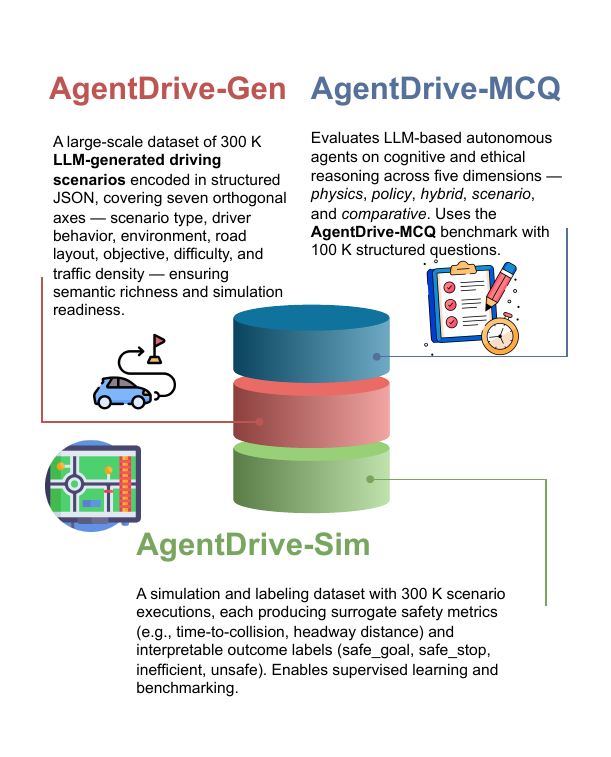} 
    \caption{Overview of the \textit{AgentDrive} benchmark suite, which comprises three complementary datasets: 
    \textit{AgentDrive-Gen} for LLM-based driving scenario generation, 
    \textit{AgentDrive-Sim} for simulation and outcome labeling, and 
    \textit{AgentDrive-MCQ} for reasoning and decision-making evaluation. }
    \label{fig:figagentdrive}
\end{figure}

To validate and further develop LLM-driven autonomous vehicles, the community is creating new benchmarks and evaluation tools to enhance their development. One notable effort is Bench2ADVLM, a closed-loop evaluation framework introduced by Zhang \textit{et al.} \cite{zhang2025bench2advlm} to test vision–language AD models in interactive settings. Bench2ADVLM enables real-time closed-loop testing across simulators and physical vehicles, with a scenario generator that uncovers failure modes. Results indicate that current systems still exhibit limitations under interactive conditions. In parallel, MAPLM \cite{cao2024maplm} provides a large-scale multimodal dataset combining maps, LiDAR, panoramic images, and Q\&A annotations, highlighting the importance of domain-specific data for grounding LLMs in driving contexts. Gao \textit{et al.} \cite{gao2025foundation} survey how foundation models can generate and analyze diverse scenarios, showing their potential for evaluating and handling safety-critical cases. Despite progress, challenges persist in latency, interpretability, privacy, regulatory compliance, and robustness. Addressing these motivates the creation of \texttt{AgentDrive}, an open benchmark that exposes LLM-based agents to diverse, safety-critical, and long-horizon scenarios through systematic closed-loop evaluation.

Building on the challenges of scalable and realistic scenario generation, our work poses two key research questions for agentic AI evaluation:

\begin{tcolorbox}[
    colback=gray!10,
    colframe=black,
    arc=6pt,
    boxrule=0.7pt,
    left=2mm, right=2mm, top=1mm, bottom=1mm,
    title=Research Questions
]
\small
\textbf{RQ1:} How can Large Language Models (LLMs) be systematically leveraged to generate diverse, semantically rich, and safety-critical autonomous driving scenarios that enable large-scale training, fine-tuning, and evaluation of agentic AI systems?\\[0.4em]
\textbf{RQ2:} How can structured simulation data and reasoning-based benchmarks be unified to comprehensively assess the cognitive, ethical, and physics-grounded reasoning capabilities of LLMs in autonomous systems?\\[0.4em]
\textbf{RQ3:} To what extent can state-of-the-art LLMs—both proprietary and open-source (e.g., GPT-5, ChatGPT 4o, DeepSeek V3, Qwen3 235B, ERNIE 4.5 300B, Grok 4)—demonstrate consistent and reliable reasoning performance across multiple driving-related reasoning dimensions?
\end{tcolorbox}

In response to these research questions, we introduce \texttt{AgentDrive}, an open benchmark dataset constructed from LLM-generated scenarios for evaluating and training agentic AI models in autonomous systems, as presented in Fig. \ref{fig:figagentdrive}. \texttt{AgentDrive} is built around an end-to-end generation and evaluation pipeline that samples from a factorized scenario space, encodes each instance into a structured specification via an LLM-driven prompt, validates and executes it in simulation, computes surrogate safety metrics, and assigns interpretable outcome labels. Furthermore, we extend this framework with \texttt{AgentDrive-MCQ}, a reasoning-oriented benchmark designed to evaluate the cognitive and decision-making capabilities of LLMs. The key contributions of this work are summarized as follows:

\begin{enumerate}
  \item \textbf{Factorized Scenario Space:} We formalize the autonomous driving scenario space across seven orthogonal axes—\emph{scenario type, driver behavior, environment, road layout, objective, difficulty,} and \emph{traffic density}—to ensure comprehensive coverage of both routine and rare safety-critical situations.

\item \textbf{AgentDrive-Gen:} We developed \texttt{AgentDrive-Gen}, a large-scale benchmark comprising 300,000 LLM-generated driving scenarios. It is built upon a prompt-engineering and schema-validation framework that transforms abstract scenario tuples into valid JSON specifications. The pipeline leverages large language models to generate semantically rich, physically consistent, and simulation-ready representations suitable for both training and evaluation.

\item \textbf{AgentDrive-Sim:} Each generated scenario from \texttt{AgentDrive-Gen} is executed within a simulator to produce dynamic rollouts. Surrogate safety metrics (e.g., minimum time-to-collision and headway distance) are computed, and a rule-based labeling system assigns interpretable outcome categories (\emph{safe\_goal}, \emph{safe\_stop}, \emph{inefficient}, \emph{unsafe}). The resulting dataset, denoted \texttt{AgentDrive-Sim}, serves as a large-scale benchmark for supervised learning and performance evaluation in safety-critical driving scenarios.

\item \textbf{AgentDrive-MCQ:} We introduced \texttt{AgentDrive-MCQ}, a reasoning-focused extension of \texttt{AgentDrive} comprising 100,000 multiple-choice questions. Each question is systematically derived from structured scenarios and categorized into five reasoning styles—\emph{physics}, \emph{policy}, \emph{hybrid}, \emph{scenario}, and \emph{comparative}. This benchmark enables large-scale, fine-grained evaluation of LLM reasoning and decision-making capabilities across diverse driving conditions, complementing simulation-based performance with structured reasoning assessment.

  \item \textbf{Large-Scale LLM Evaluation:} We conduct a comprehensive evaluation of fifty leading LLMs—including GPT-5, ChatGPT 4o, Gemini 2.5 Flash, DeepSeek V3, Qwen3 235B, ERNIE 4.5 300B, Grok 4, and Mistral Medium 3.1—on the \texttt{AgentDrive-MCQ} benchmark to measure their reasoning capabilities across five dimensions: \emph{comparative, physics, policy, hybrid,} and \emph{scenario}. Results demonstrate that while proprietary frontier models lead in contextual and policy reasoning, advanced open models are rapidly closing the gap in structured and physics-grounded reasoning.
\end{enumerate}

The remainder of this paper is organized as follows. Section~\ref{sec:related_work} reviews related work on LLM-based autonomous driving benchmarks and reasoning evaluation. Section~\ref{sec:dataset_gen} details the \texttt{AgentDrive} dataset generation methodology, including scenario design, prompt engineering, simulation, and labeling processes. Section~\ref{sec:AgentDrivemcq} introduces the \texttt{AgentDrive-MCQ} benchmark, which extends \texttt{AgentDrive} into structured reasoning evaluation. Section~\ref{sec:performance} presents the reasoning performance of fifty leading large language models on \texttt{AgentDrive-MCQ} across multiple reasoning dimensions. Finally, Section~\ref{sec:conclusion} concludes the paper and outlines future research directions.

\begin{table*}[t]
\scriptsize
\centering
\caption{Comparison of related works on LLM- and VLM-based autonomous driving benchmarks and evaluation.}
\renewcommand{\arraystretch}{0.7}
\rowcolors{2}{white}{cyanblue!70} 
\label{tab:related_work_comparison}
\begin{tabular}{p{0.08\linewidth} p{0.05\linewidth} p{0.21\linewidth} p{0.25\linewidth} p{0.31\linewidth}}
\toprule
\rowcolor{gray!35}\textbf{Work} & \textbf{Year} & \textbf{Focus / Contribution} & \textbf{Key Features} & \textbf{Limitations} \\
\midrule
\textit{LaMPilot} \cite{ma2024lampilot} & 2024 & Instruction-following AD with LLM-generated driving code & Benchmark and framework translating text to executable primitives & Limited to a pre-defined action space and code translation; does not generate structured simulation scenarios or reasoning benchmarks \\ 
\midrule
Tang \textit{et al.} \cite{tang2024test} & 2024 & Driving theory evaluation of LLMs & 500+ MCQs; comparison of GPT-3.5, GPT-4, etc. & Tests static theoretical knowledge only; lacks simulation-grounded tasks, generative scenario diversity, and reasoning integration \\ 
\midrule
\textit{V2V-LLM} \cite{chiu2025v2v} & 2025 & Cooperative vehicle-to-vehicle reasoning with multimodal LLMs & Shared perception fusion; V2V-QA dataset & Focused on connected vehicles and perception fusion; does not provide large-scale generative benchmarks or unified reasoning evaluation \\ 
\midrule
Lebioda \textit{et al.} \cite{lebioda2025requirements} & 2025 & LLMs for automotive requirement-to-code translation & Case study: natural language $\rightarrow$ CARLA configuration & Semi-automated with frequent human correction; not designed for large-scale generative or reasoning-driven evaluation \\ 
\midrule
\textit{AGENTS-LLM} \cite{yao2025agents} & 2025 & Agentic LLM for scenario augmentation & Incremental modification of real traffic logs; realistic rare cases & Relies on existing driving data; lacks fully generative, simulation-based, or reasoning-oriented evaluation \\ 
\midrule
Zhou \textit{et al.} \cite{zhou2025benchmarking} & 2025 & Benchmarking LLMs on motorway driving scenarios & Text-based scenario understanding; comparison of six LLMs & Restricted to motorway conditions and textual QA; does not incorporate generative simulation or diverse reasoning styles \\ 
\midrule
\textit{STSBench} \cite{fruhwirth2025stsbench} & 2025 & Spatio-temporal reasoning benchmark & Multi-camera nuScenes QA; 43 scenario types, 971 MCQs & Focused on visual QA; not simulation-grounded and lacks LLM-generated structured scenarios or agentic reasoning coverage \\ 
\midrule
\textit{AD$^2$-Bench} \cite{wei2025ad} & 2025 & Hierarchical CoT benchmark in adverse conditions & 5.4k annotated reasoning chains; step-by-step evaluation & Limited to perception and reasoning under predefined conditions; no generative scenario modeling or simulation-based evaluation \\ 
\midrule
Pei \textit{et al.} \cite{pei2025methodology} & 2025 & Theory + hazard perception benchmark & 700 MCQs, 54 hazard videos; real exam format & Focused on driver testing and hazard perception; lacks generative scenarios, simulation rollouts, and reasoning diversity \\ 
\midrule
\textit{DriveBench} \cite{xie2025vlms} & 2025 & Evaluation of Vision-Language Models (VLMs) for reliability and grounding & \textit{DriveBench} (VLM): 19,200 frames, 20,498 QA pairs across perception, prediction, planning, and behavior; 12 VLMs tested under 17 conditions (clean, corrupted, text-only); refined robustness metrics & Focuses on VLM robustness and reliability; not a generative benchmark—uses fixed datasets; emphasizes perception/prediction rather than textual scenario generation or reasoning diversity \\ 
\midrule
\textbf{\texttt{AgentDrive (ours)}} & 2025 & Unified benchmark suite for generative simulation and reasoning-driven evaluation of agentic AI models & 
\texttt{AgentDrive}: 300K LLM-generated, simulation-ready driving scenarios across seven orthogonal axes (scenario type, driver behavior, environment, road layout, objective, difficulty, traffic density); structured JSON schema, surrogate safety metrics, and rule-based outcome labeling. 
\texttt{AgentDrive-MCQ}: 100K reasoning questions covering five dimensions (\emph{physics}, \emph{policy}, \emph{hybrid}, \emph{scenario}, \emph{comparative}); large-scale evaluation of 50 LLMs with open-source scripts for training and assessment. &
Currently limited to text-based reasoning where visual contexts are represented as textual descriptions; multimodal (visual–language) LLMs to be integrated in future work. Unlike prior works, it introduces the first fully generative, simulation-grounded, and reasoning-oriented benchmark for evaluating autonomous agent cognition. \\ 
\bottomrule
\end{tabular}
\end{table*}

\begin{figure*}[htbp]
    \centering
    \includegraphics[width=1\textwidth]{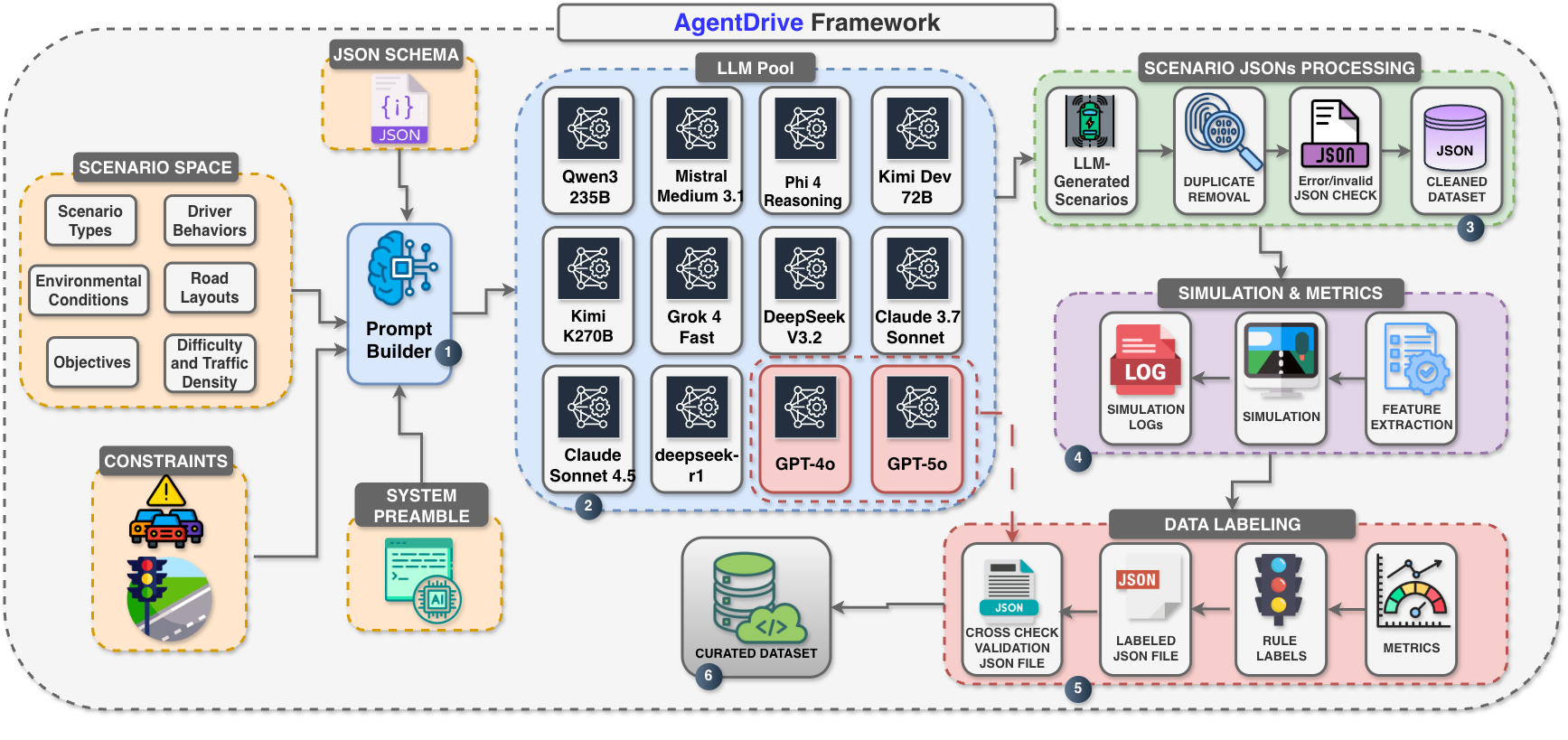} 
    \caption{\texttt{AgentDrive} Framework: End-to-End Pipeline for LLM-Generated Autonomous Driving Scenarios.}
    \label{fig:fig1}
\end{figure*}

\section{Related Work}
\label{sec:related_work}

Recent research has explored a wide range of approaches to integrating large language models into autonomous driving systems. These works span from instruction-following frameworks and cooperative driving systems to benchmarks for scenario understanding, theory testing, and automated scenario generation. For clarity, we group the discussion thematically into five main categories: (i) LLM-augmented autonomous driving agents, (ii) LLM-based scenario and code generation, (iii) benchmarks for driving scenario reasoning, (iv) driving theory tests for LLMs, and (v) our positioning of \texttt{AgentDrive} in this landscape. A comparative overview of these works, including their focus, key features, and limitations, is provided in Table~\ref{tab:related_work_comparison}.

\subsection{LLM-Augmented Autonomous Driving Agents}
Ma \textit{et al.} \cite{ma2024lampilot} introduced LaMPilot, a framework that integrates LLMs into autonomous driving systems to interpret high-level instructions and translate them into low-level driving code using predefined primitives. Alongside the framework, they released LaMPilot-Bench, a benchmark to evaluate how well LLM-based agents follow such instructions across various scenarios. Their results showed that large models can generalize across multiple driving tasks, highlighting the potential of LLMs for AD instruction-following. In the multi-vehicle context, Chiu \textit{et al.} \cite{chiu2025v2v} proposed V2V-LLM, which leverages multimodal LLMs for vehicle-to-vehicle cooperative driving. By merging perception data from multiple connected cars, V2V-LLM enables reasoning about occluded hazards and planning joint maneuvers. Their evaluations demonstrated that the approach outperforms traditional early-fusion baselines in grounding objects and supporting collaborative driving strategies.

\subsection{LLM-Based Scenario and Code Generation}
Beyond direct control, LLMs have also been used to generate simulation content. Lebioda \textit{et al.} \cite{lebioda2025requirements} studied whether requirements in natural language can be transformed into configuration code for driving simulators. Using a case study in CARLA, they demonstrated that while LLMs can generate functional code, human intervention is still required to address errors and omissions, underscoring the limitations of current models in automated scenario coding. In contrast, Yao \textit{et al.} \cite{yao2025agents} introduced AGENTS-LLM, a framework for augmenting real-world traffic scenarios with LLM-driven modifications. Their agentic approach allows incremental scenario editing (e.g., introducing a red-light violation) while maintaining physical plausibility. Human evaluations confirmed that AGENTS-LLM generates realistic yet rare edge cases, providing scalable stress-testing for autonomous planners.

\subsection{Benchmarks for Driving Scenario Reasoning}
Zhou \textit{et al.} benchmarked six LLMs on motorway driving scenario understanding, focusing on their ability to interpret textual descriptions of traffic situations \cite{zhou2025benchmarking}. Their results offer insights into the relative strengths of LLMs for functional driving scenario reasoning. Fruhwirth-Reisinger \textit{et al.} developed STSBench, a spatio-temporal benchmark derived from nuScenes that evaluates multimodal LLMs on reasoning across time and multiple views \cite{fruhwirth2025stsbench}. Covering 43 scenario types with nearly 1,000 human-verified multiple-choice questions, STSBench revealed that state-of-the-art vision-language models still struggle with spatio-temporal reasoning in traffic. Similarly, Wei \textit{et al.} presented AD$^2$-Bench, a benchmark targeting adverse driving conditions such as dense traffic, fog, and nighttime environments \cite{wei2025ad}. With more than 5,000 annotated reasoning chains, AD$^2$-Bench highlights the difficulty multimodal current LLMs face in step-by-step reasoning under safety-critical conditions.

\subsection{Driving Theory Tests for LLMs}
Several works have drawn inspiration from human licensing tests to evaluate LLMs’ knowledge of driving rules. Tang \textit{et al.} tested multiple LLMs on driving theory questions, reporting that only GPT-4 consistently exceeded the passing threshold, while other models lacked sufficient domain knowledge \cite{tang2024test}. Pei \textit{et al.} expanded this approach by creating a benchmark of over 700 multiple-choice questions and 54 hazard perception videos \cite{pei2025methodology}. Their findings showed that while GPT-4 performed strongly on theoretical questions, no model successfully passed the hazard perception component, revealing current limitations in dynamic traffic understanding.

\subsection{\texttt{AgentDrive} in Context}
Recent research has demonstrated significant progress in applying large language models to autonomous driving, spanning instruction following, cooperative reasoning, requirement translation, and domain-specific benchmarking. However, existing studies typically address isolated components of the autonomy pipeline—such as perception, code synthesis, or theoretical reasoning—without providing a unified and generative framework that integrates scenario creation, simulation, and reasoning evaluation. 

\texttt{AgentDrive} advances this landscape by introducing a fully generative, simulation-grounded, and reasoning-oriented benchmark that bridges these dimensions. It defines a formal compositional scenario space and employs LLM-driven prompt engineering to generate 300K structured driving scenarios validated through a JSON schema. Each scenario is grounded in high-fidelity simulation rollouts, where surrogate safety metrics are computed, and rule-based labeling provides interpretable outcome categories. In addition, entropy-maximized sampling ensures balanced coverage across scenario dimensions, including rare safety-critical conditions. 

Beyond simulation, \texttt{AgentDrive-MCQ} extends the benchmark into the reasoning domain with 100K multiple-choice questions that assess LLM understanding across physics, policy, hybrid, scenario, and comparative reasoning styles. This unified framework—combining generative scenario synthesis with structured reasoning evaluation—positions \texttt{AgentDrive} as the first principled benchmark for assessing the robustness, safety, and generalization capabilities of agentic AI systems in autonomous driving.

\section{\texttt{AgentDrive}: Dataset Generation Methodology}
\label{sec:dataset_gen}

This section presents our proposed pipeline for generating, simulating, and labeling autonomous driving scenarios. The methodology is structured into six key components: (i) definition of the scenario space, (ii) LLM-driven specification, (iii) simulation rollout generation, (iv) computation of surrogate safety metrics, (v) rule-based labeling, and (vi) construction of the data set. Each component is described in detail with a mathematical formalization that clarifies the problem setting and highlights the reproducibility of our framework. The primary goal is to transform abstract scenario descriptions into structured, simulation-ready inputs that yield diverse, safety-critical driving trajectories. Table~\ref{tab:notation} presents the key symbols, their corresponding methodological blocks, and the concise definitions used throughout the dataset generation pipeline.

Figure \ref{fig:fig1} illustrates the complete workflow of \texttt{AgentDrive}, which transforms abstract scenario specifications into curated, simulation-ready datasets. The process begins with a factorized scenario space that encodes driver behaviors, road layouts, environmental conditions, objectives, difficulty levels, and traffic density. Structured prompts are then built and passed to a diverse pool of large language models (LLMs), which generate candidate scenario JSONs. These outputs undergo schema validation, error checks, duplicate removal, and cross-model consistency verification before being added to the cleaned dataset. Validated scenarios are executed in simulation, during which logs and surrogate safety metrics are collected. This is followed by rule-based labeling to assign interpretable outcome categories such as safe stop, safe goal, unsafe, or inefficient. The final curated dataset provides a diverse, safety-critical benchmark that supports the training and evaluation of agentic AI systems for autonomous driving.

\subsection{Scenario Space Definition}
The foundation of reliable autonomous driving evaluation lies in constructing a representative and diverse scenario space. To formalize this, we define a scenario as a tuple:
\begin{equation}
s = (t, b, e, r, o, d, q),
\label{eq:scenario_tuple}
\end{equation}
where $t \in \mathcal{T}$ represents the scenario type (e.g., lane change, intersection crossing, merging), $b \in \mathcal{B}$ denotes the driver behavior (e.g., compliant, distracted, or aggressive), $e \in \mathcal{E}$ captures environmental conditions (e.g., rain, fog, or low visibility), $r \in \mathcal{R}$ defines the road layout (e.g., straight highway, urban intersection, or roundabout), $o \in \mathcal{O}$ specifies the ego-vehicle objective (e.g., safe navigation, overtaking, or emergency stop), $d \in \mathcal{D}$ quantifies the difficulty level (e.g., easy, moderate, or complex), and $q \in \mathcal{Q}$ encodes traffic density (e.g., sparse, medium, or congested). 

The overall scenario space is expressed as the Cartesian product of these seven axes:
\begin{equation}
\mathcal{S} = \mathcal{T} \times \mathcal{B} \times \mathcal{E} \times \mathcal{R} \times \mathcal{O} \times \mathcal{D} \times \mathcal{Q}.
\label{eq:scenario_space}
\end{equation}

This formalization provides a structured and extensible framework for representing the heterogeneity of real-world driving contexts. By sampling along each axis, we ensure broad coverage of both nominal and non-nominal conditions. Crucially, this includes rare but critical safety scenarios, such as malfunctioning traffic signals, sensor-blinding weather, or adversarial driver behaviors. These edge cases, though infrequent in naturalistic datasets, are disproportionately influential in assessing system robustness and generalization. 

Moreover, the factorized design of $\mathcal{S}$ enables systematic stress-testing across dimensions. For example, combining aggressive driver behavior ($b$) with high traffic density ($q$) under adverse weather ($e$) can generate compound scenarios that expose the limitations of perception, prediction, and planning modules. Hence, the scenario space not only supports comprehensive coverage but also facilitates controlled experimentation to benchmark agentic AI systems driven by LLM-generated scenarios.

\begin{table}[t]
\renewcommand{\arraystretch}{0.7} 
\scriptsize
\caption{Notation used throughout the dataset generation methodology.}
\rowcolors{2}{white}{cyanblue!70} 
\label{tab:notation}
\begin{tabular}{llp{3.8cm}}
\toprule
\rowcolor{gray!35}Symbol & Block & Definition \\
\midrule
$s$ & Scenario Space & Scenario tuple $s=(t,b,e,r,o,d,q)$ \\
$t\in\mathcal{T}$ & Scenario Space & Scenario type (e.g., lane change, intersection) \\
$b\in\mathcal{B}$ & Scenario Space & Driver behavior (e.g., compliant, aggressive) \\
$e\in\mathcal{E}$ & Scenario Space & Environmental conditions (e.g., rain, fog) \\
$r\in\mathcal{R}$ & Scenario Space & Road layout/topology \\
$o\in\mathcal{O}$ & Scenario Space & Ego objective (e.g., overtake, safe stop) \\
$d\in\mathcal{D}$ & Scenario Space & Difficulty level $\{\text{easy}, \text{medium}, \text{hard}\}$ \\
$q\in\mathcal{Q}$ & Scenario Space & Traffic density $\{\text{low}, \text{medium}, \text{high}\}$ \\
$\mathcal{S}$ & Scenario Space & Cartesian product $\mathcal{T}\times\mathcal{B}\times\mathcal{E}\times\mathcal{R}\times\mathcal{O}\times\mathcal{D}\times\mathcal{Q}$ \\
$H(d)$ & Scenario Space & Difficulty-specific numeric constraints (e.g., min\_ttc, max\_accel) \\
\midrule
$P(s,H(d))$ & LLM Spec & Structured prompt combining $f(s)$ with $H(d)$ \\
$f(s)$ & LLM Spec & Natural-language encoding of the scenario tuple $s$ \\
$\mathcal{G}_\phi$ & LLM Spec & LLM generator with parameters $\phi$ \\
$\hat{\jmath}$ & LLM Spec & Raw JSON scenario generated by $\mathcal{G}_\phi$ \\
$\mathcal{C}$ & LLM Spec & JSON schema / structural constraints \\
$\Pi$ & LLM Spec & Post-processing/repair module enforcing $\mathcal{C}$ \\
$R$ & LLM Spec & Maximum repair/retry attempts for $\Pi$ \\
$\hat{\jmath}^{*}$ & LLM Spec & Repaired/validated JSON after $\Pi$ \\
\midrule
$\mathcal{M}$ & Simulation & Simulator configured by $\hat{\jmath}^{*}$ \\
$\tau=\{x_t\}_{t=0}^{T-1}$ & Simulation & Rollout (state trajectory) of horizon $T$ \\
$T$ & Simulation & Number of discrete time steps (\texttt{duration\_steps}) \\
$x_t\in\mathbb{R}^d$ & Simulation & State vector at time $t$ (ego, traffic, signals, environment) \\
$\Delta t$ & Simulation & Time step, $\Delta t=1/\texttt{policy\_frequency}$ \\
\midrule
$\text{TTC}(t)$ & Metrics & Instantaneous time-to-collision at time $t$ \\
$\mathrm{TTC}_{\min}$ & Metrics & Episode-level minimum TTC, $\min_t \text{TTC}(t)$ \\
$x_E(t), x_V(t)$ & Metrics & Longitudinal positions of ego $E$ and lead vehicle $V$ \\
$v_E(t), v_V(t)$ & Metrics & Velocities of ego $E$ and lead vehicle $V$ \\
$\theta_{\text{unsafe}}$ & Metrics & Unsafe TTC threshold (e.g., $0.5\,\text{s}$) \\
$\theta_{\text{near}}$ & Metrics & Near-miss TTC threshold (e.g., $1.0\,\text{s}$) \\
\midrule
\texttt{collision} & Labeling & Detected crash event (boolean) \\
\texttt{red\_violation} & Labeling & Crossed stopline on red (boolean) \\
\texttt{stopped\_on\_red} & Labeling & Full stop before stopline on red (boolean) \\
\texttt{crossed\_green} & Labeling & Crossed stopline on green (boolean) \\
\texttt{Y} & Labeling & Outcome label $\in\{\texttt{unsafe}, \texttt{safe\_goal}, \texttt{safe\_stop}, \newline \texttt{inefficient}\}$ \\
$\ell$ & Labeling & Assigned categorical label for a rollout \\
\midrule
$\mathcal{D}$ & Dataset & Final dataset $\{(s_i,\hat{\jmath}_i,\tau_i,\ell_i)\}_{i=1}^N$ \\
$N$ & Dataset & Number of generated scenarios/entries \\
$\mathcal{H}(\cdot)$ & Dataset & Entropy used to encourage axis-wise coverage \\
$\{axis_i\}$ & Dataset & Sequence of sampled values along a given axis ($i=1..N$) \\
\texttt{policy\_frequency} & Params & Control frequency (Hz) determining $\Delta t$ \\
\texttt{duration\_steps} & Params & Discrete horizon length for each scenario \\
\bottomrule
\end{tabular}
\end{table}

\subsection{Scenario Taxonomy and Diversity}
The richness of the scenario space $\mathcal{S}$ (Eq.~\ref{eq:scenario_space}) emerges from its decomposition into multiple orthogonal axes, each capturing a distinct facet of the driving task. This factorized view not only supports structured exploration of safety-critical situations but also enables controlled variation along individual dimensions. Below, we outline the key axes and their corresponding taxonomies.

\subsubsection{Scenario Types.} 
We define a comprehensive set of driving situations that span both routine and edge-case conditions. Representative examples include \textit{red-light dilemma zones}, \textit{yield/priority confusion}, \textit{highway merging at high speed}, \textit{icy bridge segments}, and \textit{pedestrian jaywalking}. The taxonomy explicitly incorporates vulnerable road users (VRUs), environmental hazards, and mechanical failures (e.g., brake loss, tire blowouts). By explicitly encoding corner cases, the taxonomy ensures systematic stress-testing of autonomous agents in situations rarely observed in naturalistic driving logs but disproportionately critical for safety validation.

\subsubsection{Driver Behaviors.} 
To capture the heterogeneity of human driving styles, we define behavioral profiles ranging from \textit{calm, cooperative, or rule-following} to \textit{aggressive, impatient, distracted, or impaired}. We also account for role-specific driving modes, such as \textit{taxi-style stop-and-go}, \textit{delivery routing under time pressure}, and \textit{convoy following}. This axis ensures that autonomous systems are evaluated not only against compliant traffic participants but also against unpredictable, adversarial, and non-standard agents, reflecting the variability of real-world traffic ecosystems.

\subsubsection{Environmental Conditions.} 
Real-world conditions vary widely across geography, climate, and time of day. Our taxonomy spans clear daytime driving, low-visibility situations such as \textit{foggy night driving}, and extreme weather events including \textit{hailstorms}, \textit{sandstorms}, and \textit{blinding low-sun glare}. We also incorporate urban-specific challenges such as \textit{neon light glare at night}, which degrade visual perception systems. These variations directly reflect the operational design domains (ODDs) outlined by regulatory bodies and expose the fragility of perception and control pipelines under degraded sensing conditions.

\subsubsection{Road Layouts.} 
Road geometry constitutes another axis of complexity. We include highways, interchanges, roundabouts, rural roads, mountain passes, tunnels, and bridges with varying constraints such as \textit{narrow lanes}, \textit{limited visibility}, or \textit{sharp curves}. By systematically varying these topologies, the framework probes the limits of map-based localization, lane-level reasoning, and motion planning under non-trivial geometric constraints.

\subsubsection{Objectives.} 
The ego vehicle's intended task defines the contextual goal for each scenario. Objectives range from regulatory compliance (e.g., \textit{obey red light}, \textit{respect pedestrian crossings}, \textit{use turn signals}) to maneuvering (e.g., \textit{merge smoothly}, \textit{overtake safely}, \textit{execute U-turns}). Defensive objectives, such as \textit{maintaining safe headway} or \textit{avoiding collision}, are also integrated, aligning with safety-critical evaluation metrics. By conditioning on objectives, scenarios can be tailored to assess specific capabilities of planning and decision-making modules.

\begin{figure*}[t]
    \centering
    \begin{subfigure}[b]{0.32\textwidth}
        \centering
        \includegraphics[width=0.8\textwidth]{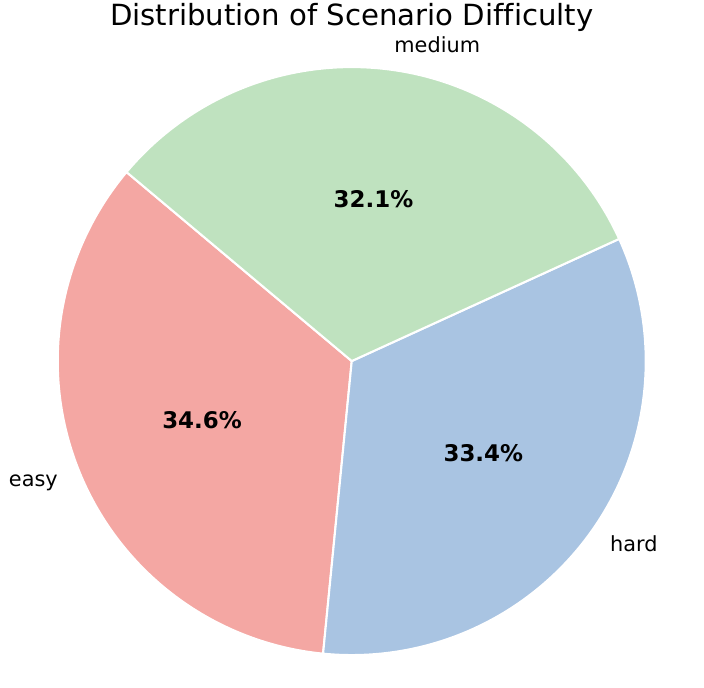}
        \caption{Scenario Difficulty Distribution}
    \end{subfigure}
    \hfill
    \begin{subfigure}[b]{0.32\textwidth}
        \centering
        \includegraphics[width=\textwidth]{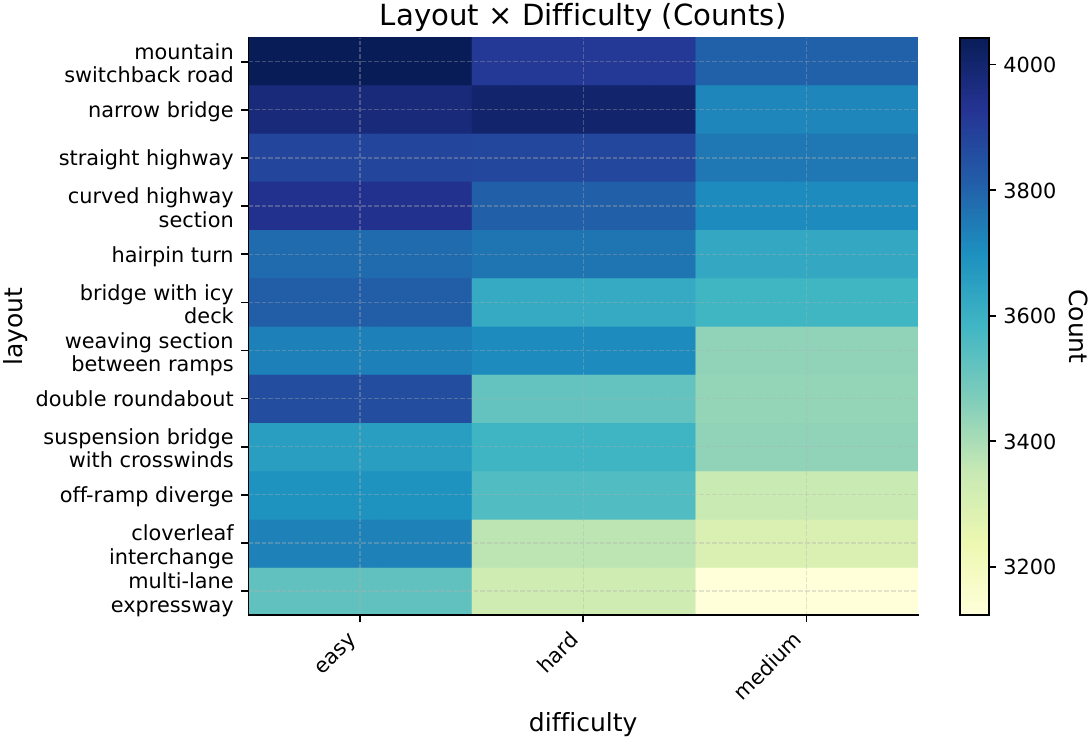}
        \caption{Layout vs. Difficulty Heatmap}
    \end{subfigure}
    \hfill
    \begin{subfigure}[b]{0.32\textwidth}
        \centering
        \includegraphics[width=\textwidth]{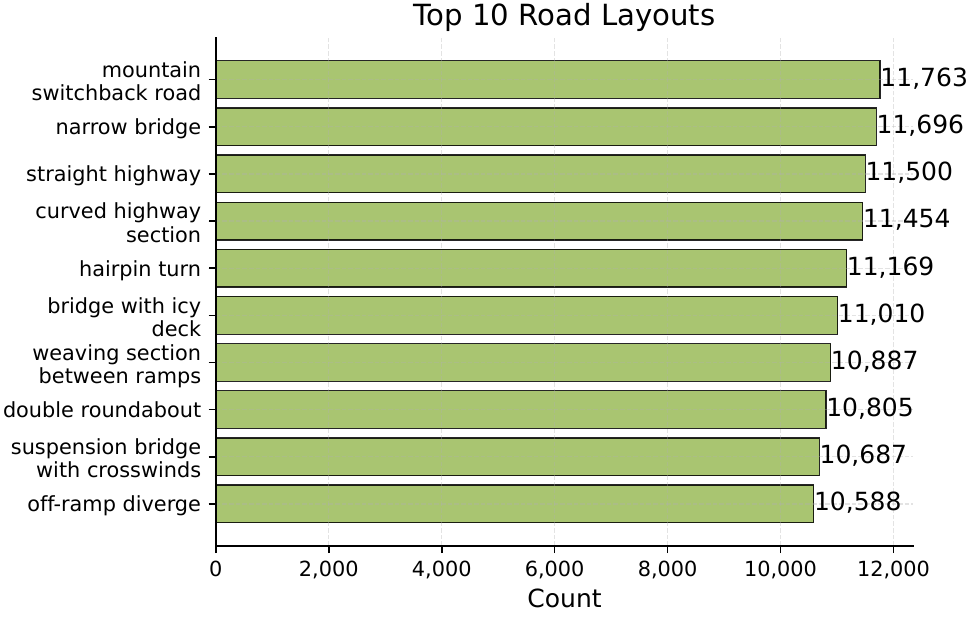}
        \caption{Top 10 Road Layouts}
    \end{subfigure}

    \vspace{0.5cm}

    \begin{subfigure}[b]{0.32\textwidth}
        \centering
        \includegraphics[width=\textwidth]{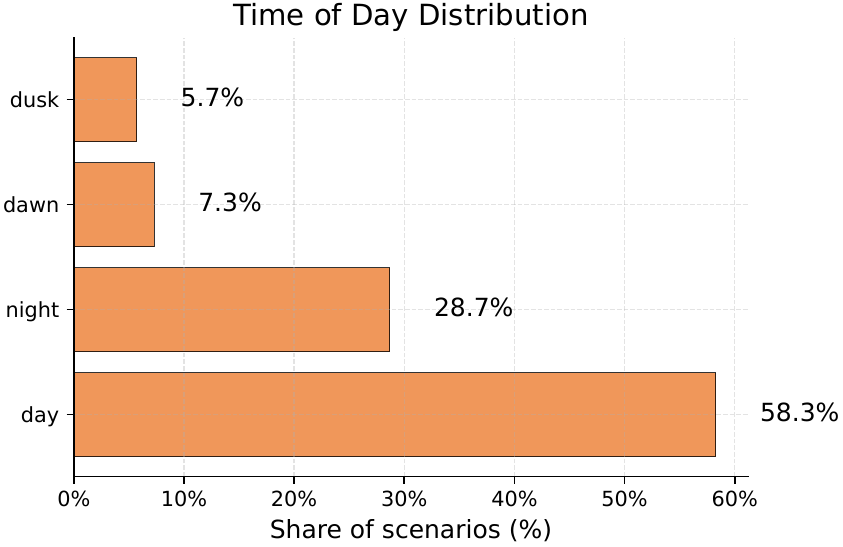}
        \caption{Time of Day Distribution}
    \end{subfigure}
    \hfill
    \begin{subfigure}[b]{0.32\textwidth}
        \centering
        \includegraphics[width=\textwidth]{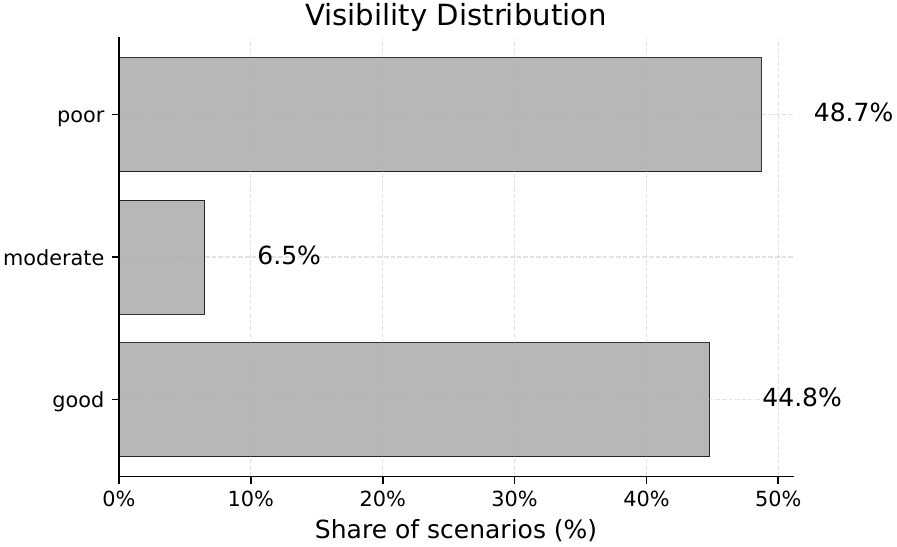}
        \caption{Visibility Conditions}
    \end{subfigure}
    \hfill
    \begin{subfigure}[b]{0.32\textwidth}
        \centering
        \includegraphics[width=\textwidth]{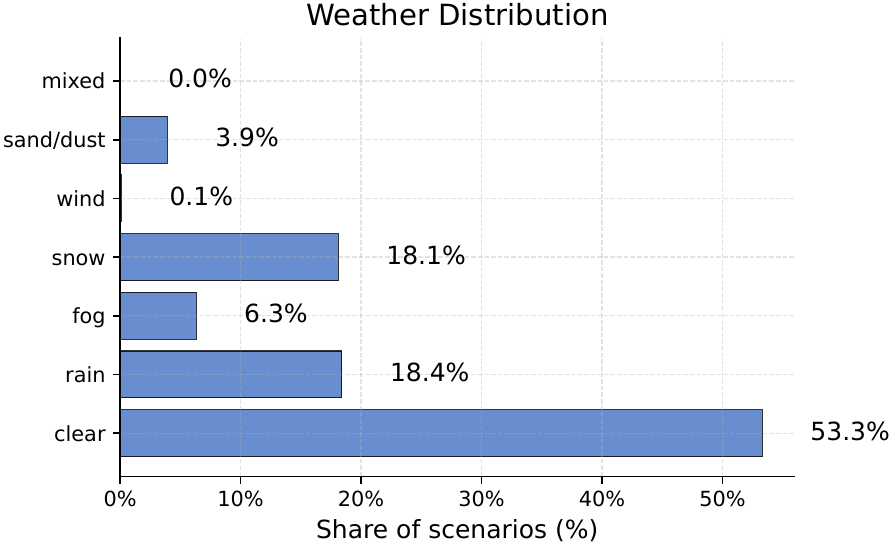}
        \caption{Weather Distribution}
    \end{subfigure}

    \caption{Distributions and layout statistics across the AgentDrive scenario dataset, showing difficulty levels, layout correlations, temporal and environmental characteristics.}
    \label{fig:AgentDrive_dataset_overview}
\end{figure*}

\subsubsection{Difficulty and Traffic Density.} 
To control scenario challenge in a principled way, we vary difficulty levels $\{\text{easy}, \text{medium}, \text{hard}\}$ and traffic densities $\{\text{low}, \text{medium}, \text{high}\}$. Each difficulty level is mapped to structured constraints $H(d)$ that define numerical thresholds for safety-critical quantities such as minimum time-to-collision (TTC) and maximum admissible acceleration:
\begin{align}
H(\text{easy})   &= \{\text{min\_ttc}=4.0~s,\ \text{max\_accel}=2.5~m/s^2\},\\
H(\text{medium}) &= \{\text{min\_ttc}=2.5~s,\ \text{max\_accel}=4.0~m/s^2\},\\
H(\text{hard})   &= \{\text{min\_ttc}=1.5~s,\ \text{max\_accel}=6.0~m/s^2\}.
\end{align}
These constraints serve as structured hints to the LLM during scenario generation, ensuring that the resulting instances conform to quantifiable levels of risk and challenge.

\medskip
By combining these axes, \texttt{AgentDrive} systematically generates scenarios that balance routine coverage with exposure to rare, safety-critical events. This design positions \texttt{AgentDrive} not only as a diverse dataset but also as a principled benchmark for robustness, safety, and generalization in autonomous systems powered by agentic AI.

\subsection{Scenario Specification via LLM}
Once a scenario tuple is sampled, the next challenge is to transform its abstract representation into a structured, semantically rich specification suitable for simulation. This transformation requires bridging symbolic scenario definitions with executable formats that respect both linguistic expressiveness and physical realism.

Formally, for each sampled scenario $s$, we construct a prompt:
\begin{equation}
P(s, H(d)) = f(s) \cup H(d),
\label{eq:prompt}
\end{equation}
where $f(s)$ encodes the axis tuple into a natural language description, and $H(d)$ provides difficulty-specific numerical hints (e.g., minimum time-to-collision thresholds, maximum acceleration bounds). The prompt thus combines contextual semantics with quantitative constraints, guiding the generative process.

A large language model (LLM) $\mathcal{G}_\phi$ then produces a structured JSON representation:
\begin{equation}
\hat{\jmath} = \mathcal{G}_\phi\big(P(s, H(d))\big),
\label{eq:llm_generation}
\end{equation}
which must conform to a schema $\mathcal{C}$. Generations that violate $\mathcal{C}$ are flagged as invalid and corrected by a post-processing module $\Pi$, which retries up to $R$ attempts:
\begin{equation}
\hat{\jmath}^{*} = \Pi(\hat{\jmath}, \mathcal{C}, R).
\end{equation}

To operationalize the process, we define a structured prompt consisting of three components:  
\begin{tcolorbox}[
    colback=gray!10,
    colframe=black,
    arc=6pt,
    boxrule=0.7pt,
    left=2mm, right=2mm, top=1mm, bottom=1mm
]
\small
\begin{itemize}[leftmargin=*, itemsep=2pt]
    \item \textbf{System preamble:} 
    \begin{quote}
        \texttt{You are a driving scenario planner for a highway simulator. Return a SINGLE valid JSON object that exactly matches the schema. Do not include comments or explanations—JSON ONLY.}
    \end{quote}

    \item \textbf{Scenario description $f(s)$ and difficulty hints $H(d)$}, e.g., “Scenario type: red-light dilemma, driver: aggressive, environment: foggy night, road: urban intersection, objective: avoid collision, difficulty: hard, traffic density: high. Constraints: min\_ttc=1.5s, max\_accel=6.0 m/s².”

    \item \textbf{Constraints:}
    \begin{quote}
        \texttt{Respect physics: |accel| $\leq$ 6 m/s², v\_mps $\geq$ 0, lanes are 0-indexed.}\\
        \texttt{If traffic\_light present: ego spawn.x $\geq$ 30 m before stopline\_x.}\\
        \texttt{duration\_steps $\geq$ red\_steps + green\_steps.}\\
        \texttt{Include seed, policy\_frequency, and all required fields.}
    \end{quote}
\end{itemize}
\end{tcolorbox}

The generated JSON must conform to the schema $\mathcal{C}$, summarized in Table~\ref{tab:json_schema}.

\begin{table}
\renewcommand{\arraystretch}{0.7} 

\centering
\caption{Schema for LLM-generated scenarios .}
\renewcommand{\arraystretch}{0.7}
\rowcolors{2}{white}{cyanblue!70} 
\label{tab:json_schema}
\begin{tabular}{p{0.25\linewidth} p{0.65\linewidth}}
\toprule
\rowcolor{gray!35}\textbf{Field} & \textbf{Description / Constraints} \\
\midrule
\texttt{name} & Short slug, e.g., \texttt{RedLight\_Merge\_v1} \\
\texttt{seed} & Non-negative integer \\
\texttt{duration\_steps} & Integer $\geq 60$ (0.1s per step at policy\_frequency=10) \\
\texttt{road} & \{\texttt{lanes} $\geq 2$, \texttt{speed\_limit\_kph} $\in [60,140]$\} \\
\texttt{traffic\_light} & Stopline position, red/green durations \\
\texttt{environment} & Weather, time of day, visibility \\
\texttt{layout} & Text description of road geometry \\
\texttt{objective} & Ego-vehicle task (e.g., avoid collision) \\
\texttt{difficulty} & \texttt{easy|medium|hard} \\
\texttt{traffic\_density} & \texttt{low|medium|high} \\
\texttt{ego.spawn} & \{x, y, v\_mps $\geq 0$\}, plus goal description \\
\texttt{traffic[]} & List of agents (type, behavior, lane, spawn, velocity) \\
\texttt{events[]} & Optional timed events (e.g., sudden brake, cut-in) \\
\texttt{metrics[]} & Evaluation metrics (TTC, headway, collisions) \\
\bottomrule
\end{tabular}
\end{table}

Figure~\ref{fig:AgentDrive_dataset_overview} presents an overview of the AgentDrive scenario dataset, which comprises around 300,000 samples, highlighting its diversity across difficulty levels, road layouts, and environmental conditions. Subfigure~(a) shows a well-balanced distribution of scenario difficulty, with roughly one-third of the samples categorized as \textit{easy}, \textit{medium}, and \textit{hard}. Subfigure~(b) illustrates the correlation between layout type and difficulty, indicating that complex road structures—such as \textit{hairpin turns} and \textit{cloverleaf interchanges}—are generally associated with higher difficulty levels. Subfigure~(c) lists the ten most frequent road layouts, showcasing the dataset’s variety of geometric configurations. Subfigures~(d)–(f) capture temporal and environmental diversity, showing that most scenarios occur during \textit{daytime}, under \textit{good visibility}, and in \textit{clear weather}—conditions that ensure both realism and safety-critical coverage for evaluating autonomous driving systems.

\begin{figure}[t]
    \centering
    \begin{tikzpicture}
    \begin{axis}[
        ybar,
        bar width=10pt,
        width=0.95\columnwidth, height=8.5cm,
        ymin=0,
        ylabel={Count},
        symbolic x coords={unsafe,safe-goal,safe-stop,inefficient},
        xtick=data,
        xticklabel style={yshift=-10pt},
        enlarge x limits=0.25,
        legend style={
            at={(0.5,0.96)},        
            anchor=south,
            legend columns=-1,
            font=\small,
            draw=black,
            fill=white,
            column sep=8pt,
            rounded corners=2pt
        },
        nodes near coords,
        every node near coord/.append style={
            font=\footnotesize,
            anchor=north,
            yshift=-3pt,
            text=black
        },
        ymajorgrids,
        scaled y ticks=false,
        yticklabel style={/pgf/number format/fixed, font=\small},
    ]

    \definecolor{epcol}{HTML}{3B5599}   
    \definecolor{segcol}{HTML}{D8B137}  

    \addplot[fill=epcol!60,draw=epcol] coordinates {
        (unsafe,102300) (safe-goal,91500) (safe-stop,75900) (inefficient,30600)
    };

    \addplot[fill=segcol!60,draw=segcol] coordinates {
        (unsafe,1794739) (safe-goal,5156085) (safe-stop,1152081) (inefficient,874736)
    };

    \legend{Episode, Segment}
    \end{axis}
    \end{tikzpicture}

    \caption{Label distribution of \texttt{AgentDrive-Sim} across episode and segment levels.}
    \label{fig:AgentDrive_labels}
\end{figure}
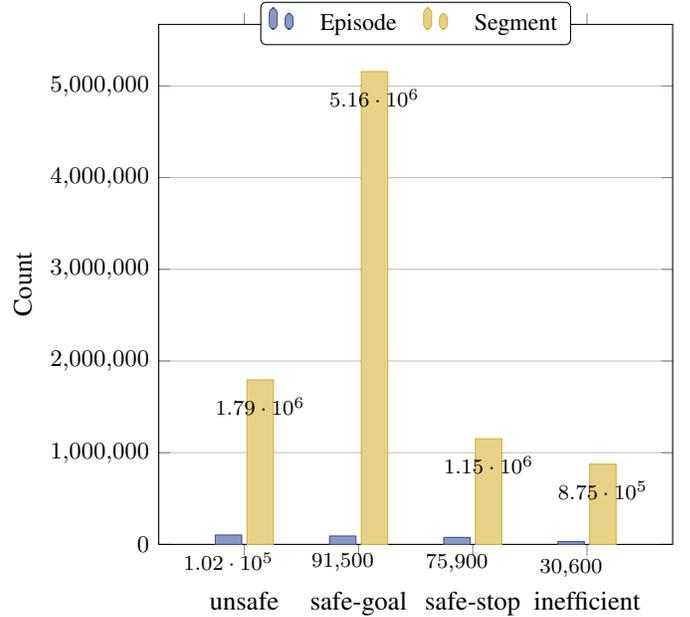

\subsection{Simulation Rollout Generation}
Once a valid JSON specification $\hat{\jmath}$ has been generated, the simulator $\mathcal{M}$ is implemented using the \texttt{highway-env} framework~\cite{highway-env} and configured to produce a trajectory (or rollout):
\begin{equation}
\tau = \{x_t\}_{t=0}^{T-1},
\label{eq:rollout}
\end{equation}
where $T$ is the horizon determined by \texttt{duration\_steps}, and each state $x_t \in \mathbb{R}^d$ encodes ego-vehicle kinematics (position, velocity, acceleration), surrounding traffic states, traffic light signals, and relevant environmental parameters. The simulation time step is governed by the policy frequency:
\begin{equation}
\Delta t = \frac{1}{\texttt{policy\_frequency}}.
\label{eq:time_step}
\end{equation}

The rollout $\tau$ captures the temporal dynamics of vehicle interactions, providing a reproducible and controllable environment for downstream analysis. The use of \texttt{highway-env}~\cite{highway-env} allows for efficient prototyping and standardized simulation of diverse driving scenarios, including highway, intersection, and merging behaviors.

\subsection{Surrogate Safety Metrics}
To evaluate safety outcomes at scale without requiring costly real-world testing, we rely on surrogate safety metrics. Among these, \emph{Time-to-Collision (TTC)} is the primary indicator of criticality. The instantaneous TTC between ego vehicle $E$ and its nearest lead vehicle $V$ is:
\begin{equation}
\text{TTC}(t) =
\begin{cases}
\dfrac{x_V(t) - x_E(t)}{v_E(t) - v_V(t)}, & v_E(t) > v_V(t) \wedge x_V(t) > x_E(t), \\
+\infty, & \text{otherwise},
\end{cases}
\label{eq:ttc}
\end{equation}
where $x_E(t)$, $x_V(t)$ are longitudinal positions, and $v_E(t)$, $v_V(t)$ are velocities of ego and lead vehicles. The episode-level critical TTC is given by:
\begin{equation}
\mathrm{TTC}_{\min} = \min_t \text{TTC}(t).
\label{eq:ttc_min}
\end{equation}

We define thresholds $\theta_{\text{unsafe}}=0.5\,s$ and $\theta_{\text{near}}=1.0\,s$, consistent with established traffic safety analysis standards. These thresholds allow us to classify events into unsafe, near-miss, and safe zones. Such surrogate metrics provide interpretable signals that map closely to real-world safety concepts, enabling automated large-scale safety assessment.

\subsection{Rule-Based Labeling}
While continuous metrics quantify risk, categorical labels improve interpretability and facilitate supervised learning. From each rollout $\tau$, we detect discrete events:
{\small
\begin{equation}
\begin{aligned}
\event{collision}            &= (\text{ego crashed}),\\
\event{red\_violation}       &= (\text{crossed stopline} \land (\text{light}=\text{Red})),\\
\event{stopped\_on\_red}     &= (\neg\,\text{crossed stopline} \land (v_E<\epsilon) \land (\text{light}=\text{Red})),\\
\event{crossed\_green}&= (\text{crossed stopline} \land (\text{light}=\text{Green})).
\end{aligned}
\end{equation}
}
These events are mapped into high-level outcome labels:
{\small
\begin{equation}
\begin{aligned}
\mathsf{Y} &= \texttt{unsafe},      && \text{if } \event{collision} \lor \event{red\_violation} \lor \mathrm{TTC}_{\min}<\theta_{\text{unsafe}},\\
\mathsf{Y} &= \texttt{safe\_goal},  && \text{if } \event{stopped\_on\_red} \land \event{crossed\_green},\\
\mathsf{Y} &= \texttt{safe\_stop},  && \text{if } \event{stopped\_on\_red} \land \lnot \event{crossed\_green},\\
\mathsf{Y} &= \texttt{inefficient}, && \text{otherwise}.
\end{aligned}
\label{eq:labeling}
\end{equation}
}
This labeling strategy strikes a balance between precision and interpretability. It assigns outcomes recognizable to human experts (e.g., \emph{unsafe}, \emph{safe stop}), thereby complementing continuous metrics with discrete ground truth categories. Fig. \ref{fig:AgentDrive_labels} presents the label distribution of \texttt{AgentDrive-Sim} across both episode and segment levels. The results show that at the segment level, safe-goal instances dominate with more than five million samples, followed by unsafe and inefficient outcomes, whereas the episode-level distribution remains more balanced across all categories. This clear disparity highlights the higher granularity and diversity of behaviors captured during segment-level analysis compared to aggregated episode-level evaluations.

\subsection{Dataset Construction}
After $N$ iterations, we obtain the benchmark dataset:
\begin{equation}
\mathcal{D} = \{(s_i, \hat{\jmath}_i, \tau_i, \ell_i)\}_{i=1}^N,
\label{eq:dataset}
\end{equation}
where each entry contains: the sampled scenario axes $s_i$, the generated JSON $\hat{\jmath}_i$, the resulting rollout $\tau_i$, and the assigned label $\ell_i$. To ensure coverage, we maximize entropy across scenario axes:
\begin{equation}
\max \sum_{axis \in \{t,b,e,r,o,d,q\}} \mathcal{H}\big(\{axis_i\}\big),
\label{eq:entropy}
\end{equation}
subject to schema compliance $\hat{\jmath}_i \in \mathcal{C}$ and valid rollouts $\tau_i$.

This entropy maximization prevents the dataset from collapsing into trivial or repetitive cases, ensuring diversity across scenario types, behaviors, environments, and road layouts. The resulting dataset thus captures both routine and rare conditions, providing a rigorous benchmark for training and evaluating autonomous driving models under varied and safety-critical contexts.

\begin{algorithm}
\caption{\texttt{AgentDrive} Scenario Generation and Labeling Pipeline}
\label{alg:pipeline}
\small
\begin{algorithmic}[1]
\State Sample scenario tuple $s \sim \mathcal{S}$
\State Construct enriched prompt $P = P(s, H(d))$
\State Generate JSON $\hat{\jmath} = \mathcal{G}_\phi(P)$
\State Validate $\hat{\jmath}$ against schema $\mathcal{C}$; repair via $\Pi$ up to $R$ retries
\State Simulate rollout $\tau = \mathcal{M}(\hat{\jmath})$
\State Compute surrogate safety metrics (e.g., $\mathrm{TTC}_{\min}$)
\State Detect events and assign categorical label $\ell$ via Eq.~\ref{eq:labeling}
\State Store $(s, \hat{\jmath}, \tau, \ell)$ into dataset $\mathcal{D}$
\end{algorithmic}
\end{algorithm}

\subsection{End-to-End Scenario Generation Pipeline}
The overall pipeline is summarized in Algorithm~\ref{alg:pipeline}. Each stage progressively refines the representation of a scenario, moving from abstract symbolic specifications to concrete simulation trajectories with interpretable safety labels. This layered transformation ensures that high-level abstractions such as \emph{“aggressive merging during rain”} can be systematically translated into structured inputs, simulated behaviors, and quantitative safety outcomes.

The process begins by sampling scenario tuples from the multidimensional space $\mathcal{S}$ (Eq.~\ref{eq:scenario_space}), guaranteeing balanced coverage across scenario types, driver behaviors, environmental conditions, and traffic densities. This systematic sampling provides exposure to both common conditions and rare, safety-critical cases. Each tuple is then encoded into a prompt enriched with difficulty-specific hints $H(d)$ (Eq.~\ref{eq:prompt}), which conditions the large language model (LLM) to generate structured specifications that remain semantically meaningful while adhering to physics and traffic safety constraints. 

The LLM produces a candidate JSON specification (Eq.~\ref{eq:llm_generation}), which is validated against the schema $\mathcal{C}$. Invalid generations are corrected by the post-processor $\Pi$, with retries up to $R$ attempts. This stage acts as a safeguard, ensuring that only semantically correct and physically plausible specifications propagate downstream.

Once a valid JSON is obtained, the simulator $\mathcal{M}$ generates a rollout trajectory $\tau$ (Eq.~\ref{eq:rollout}) that captures the temporal evolution of the ego vehicle and surrounding agents under the defined scenario. This simulation grounds abstract descriptions in dynamic, reproducible trajectories. From $\tau$, surrogate safety metrics such as $\mathrm{TTC}_{\min}$ (Eq.~\ref{eq:ttc_min}) are computed, and discrete rule-based event detection is performed. The outcome is assigned a categorical label $\ell$ (Eq.~\ref{eq:labeling})—\texttt{safe\_goal}, \texttt{safe\_stop}, \texttt{inefficient}, or \texttt{unsafe}—providing interpretable targets for benchmarking autonomous driving models.

Finally, the complete record $(s, \hat{\jmath}, \tau, \ell)$ is stored in the dataset $\mathcal{D}$ (Eq.~\ref{eq:dataset}). Repeating this process across $N$ iterations yields a diverse, large-scale benchmark that balances coverage and realism, forming the foundation of \texttt{AgentDrive}.

\begin{table}[t]
\renewcommand{\arraystretch}{0.7} 
\scriptsize
\centering
\caption{Notation used throughout the \texttt{AgentDrive-MCQ} dataset generation methodology.}
\renewcommand{\arraystretch}{0.7}
\rowcolors{2}{white}{cyanblue!70} 
\label{tab:notation2}
\begin{tabular}{llp{4.8cm}}
\toprule
\textbf{Symbol} & \textbf{Block} & \textbf{Definition} \\
\midrule
$\mathcal{S}$   & Scenario Input & Validated scenario JSON containing structured fields \\
$d$             & Description    & Natural language narrative derived from $\mathcal{S}$ \\
$m$             & Model          & Large language model used for description/MCQ generation \\
$R$             & Control        & Retry budget for error-guided synthesis attempts \\
$Q$             & MCQ Synthesis  & Question text derived from description $d$ \\
$\mathcal{O}$   & MCQ Synthesis  & Set of four candidate answer options \\
$i^*$           & MCQ Synthesis  & Index of the correct choice ($i^* \in \{A,B,C,D\}$) \\
$r$             & MCQ Synthesis  & Concise textual rationale supporting the correct answer \\
$s$             & Style Control  & Reasoning style $\in \{\texttt{physics}, \texttt{policy}, \texttt{hybrid}, \newline \texttt{scenario}, \texttt{comparative}\}$ \\
$\delta$        & Validation     & Difficulty level $\in \{\texttt{Easy},\texttt{Medium},\texttt{Hard}\}$ \\
$M_s$           & Assembly       & Final MCQ object for style $s$ linked to scenario $\mathcal{S}$ \\
$h$             & Assembly       & SHA-256 hash for persistent identifier of $M_s$ \\
\midrule
$v$             & Physics        & Ego vehicle speed (m/s or km/h) \\
$t_r$           & Physics/Policy & Driver or agent reaction time (s) \\
$a_{\text{ego}}$& Physics        & Ego vehicle deceleration capability (m/s$^2$) \\
$a_{\text{lead}}$& Physics       & Lead/cut-in vehicle deceleration (m/s$^2$) \\
$g$             & Physics/Hybrid & Computed minimum safe headway distance (m) \\
$\tau$          & Policy         & Time-gap rule (s), e.g., 2--4 seconds depending on conditions \\
$g_{\text{hyb}}$& Hybrid         & Headway including both physics-based minimum and policy margin (m) \\
$A$             & Comparative    & Candidate driving maneuver (e.g., brake, accelerate, change lane, maintain speed) \\
\bottomrule
\end{tabular}
\end{table}

\begin{figure*}[t]
\centering
\includegraphics[width=\textwidth]{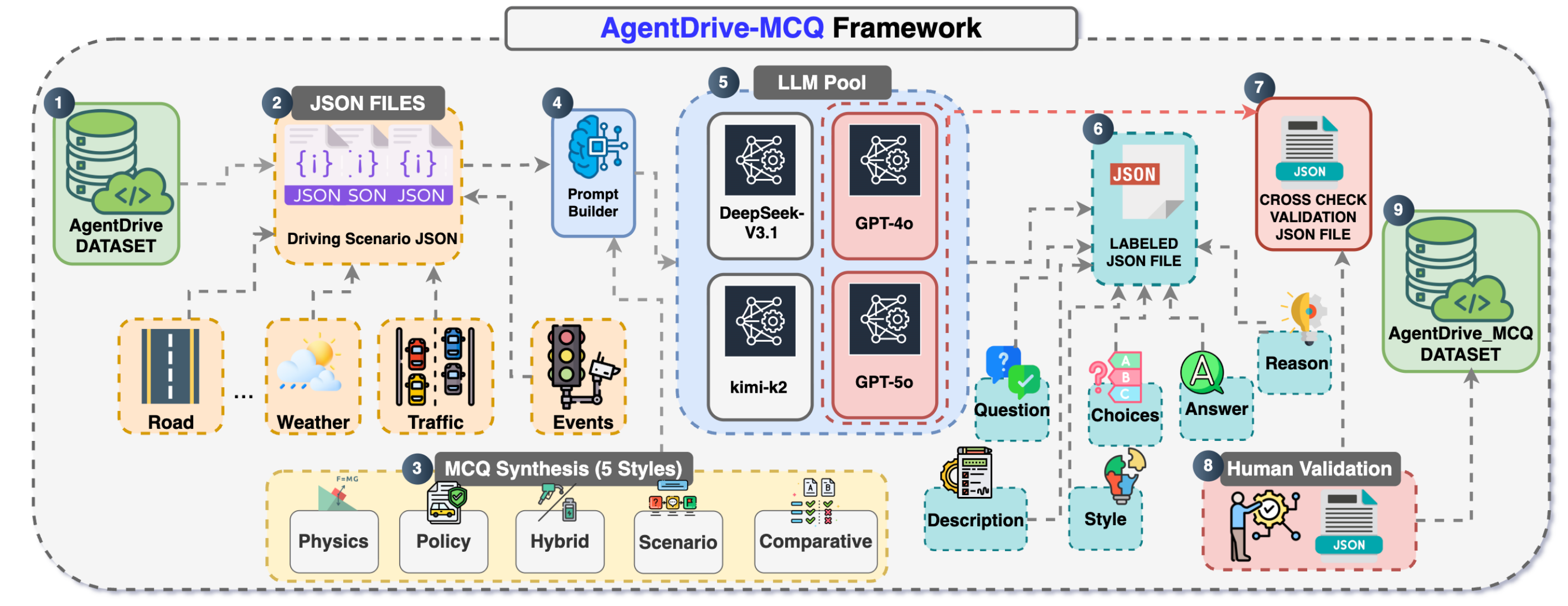}
\caption{The \texttt{AgentDrive-MCQ} framework.}
\label{fig:AgentDrive_mcq}
\end{figure*}

\begin{algorithm}[t]
\caption{\texttt{AgentDrive-MCQ} Generation}
\label{alg:AgentDrive_mcq}
\small
\KwIn{Validated scenario $\mathcal{S}$ (JSON-conformant), model $m$, retry budget $R \ge 1$}
\KwOut{Set of MCQ JSON objects $\{M_{style}\}$ linked to $\mathcal{S}$}

\tcp{Stage 1: Scenario $\rightarrow$ Description}
$d \leftarrow \textsc{GenerateDescription}(\mathcal{S}, m)$\;
\Indp
Call LLM with constrained system prompt (10--12 sentences, plain text only; include units; no lists/code)\;
Normalize whitespace; truncate to a maximum token/word budget; attach as $\mathcal{S}[\texttt{description}] \leftarrow \{\texttt{model}: m, \texttt{text}: d\}$\;
Persist augmented copy as \texttt{\textless scenario\_name\textgreater\_withdesc.json}\;
\Indm

\tcp{Stage 2: Description $\rightarrow$ MCQ Synthesis Across Five Styles}
\ForEach{$s \in \{\texttt{physics}, \texttt{policy}, \texttt{hybrid}, \texttt{scenario}, \texttt{comparative}\}$}{
  $(Q, \mathcal{O}, i^*, r) \leftarrow \textsc{MakeMCQ}(d, m, s)$ with \texttt{response\_format=json\_object}\;
  \Indp
  Validate style-specific constraints: \\
  \quad physics/hybrid $\Rightarrow$ numeric tokens in $Q$ and $\mathcal{O}$; \\
  \quad scenario $\Rightarrow$ qualitative, not all-numeric; \\
  \quad comparative $\Rightarrow$ $\geq$2 action-oriented options; \\
  Enforce general checks: $|\mathcal{O}|=4$; labels $\{A,B,C,D\}$; options distinct; $|Q|\leq 25$ words; $r\neq \varnothing$\;
  If validation fails, retry with error-guided hint up to $R$ times\;
  \Indm
}

\tcp{Stage 3: Assembly \& Persistence}
\ForEach{validated style $s$}{
  $M_s \leftarrow \langle \mathcal{S}[\texttt{name}], d, Q, \mathcal{O}, i^*, r, s \rangle$\;
  $h \leftarrow \textsc{SHA256}(d \parallel Q \parallel \mathcal{O} \parallel i^* \parallel s)$\;
  Save $M_s$ as \texttt{\textless scenario\_name\textgreater\_\{h[0:8]\}\_mcq\_\{s\}.json}\;
}
\Return $\{M_{s}\}$\;
\end{algorithm}

\section{\texttt{AgentDrive-MCQ} Benchmark}
\label{sec:AgentDrivemcq}

\subsection{Motivation}
Traditional benchmarks for autonomous driving agents have primarily emphasized perception and low-level control, including object detection, trajectory prediction, and actuation accuracy. While these tasks are essential, real-world driving introduces \emph{context-sensitive decision reasoning}, where multiple heterogeneous factors such as weather conditions, traffic signals, surrounding vehicle behavior, and road geometry must be simultaneously integrated. This dimension of reasoning is often underexplored in existing datasets, which limits their ability to evaluate agents in high-stakes, safety-critical scenarios. 

To address this gap, we introduce \texttt{AgentDrive-MCQ}, a benchmark designed to probe the reasoning and decision-making capabilities of large language models (LLMs) when deployed as agentic controllers in autonomous driving. The central innovation lies in transforming structured driving scenarios into natural language descriptions and subsequently into multiple-choice questions (MCQs) that demand hard reasoning. Unlike simple factual Q\&A, \texttt{AgentDrive-MCQ} requires agents to synthesize information across multiple scenario layers and to select strategies aligned with safety-critical objectives. 

To capture the full spectrum of driving reasoning, \texttt{AgentDrive-MCQ} spans five complementary MCQ styles: 
\begin{itemize}
  \item \textbf{Physics-based:} numerically grounded questions requiring kinematic calculations such as time-to-collision or stopping distance. 
  \item \textbf{Policy-based:} questions emphasizing traffic rules, defensive driving guidelines, and safety heuristics. 
  \item \textbf{Hybrid:} items combining physics-derived calculations with additional policy or margin-based reasoning. 
  \item \textbf{Scenario-interpretive:} qualitative reasoning about primary risks, priorities, and contextual constraints. 
  \item \textbf{Comparative/optimization:} action-selection questions evaluating the relative safety of candidate maneuvers.
\end{itemize}
This taxonomy ensures that \texttt{AgentDrive-MCQ} not only evaluates mathematical reasoning, but also policy compliance, risk interpretation, and maneuver optimization—core facets of real-world driving intelligence. Correct answers are consistently aligned with safe driving strategies, anchoring the benchmark in risk-aware reasoning.

\subsection{\texttt{AgentDrive-MCQ}: Dataset Generation Methodology}
We design a unified pipeline that consistently transforms structured scenarios into benchmark-ready MCQs across all five reasoning styles. The pipeline operates in three stages: (i) generating natural language descriptions, (ii) synthesizing reasoning-intensive MCQs in each style, and (iii) assembling validated MCQ objects with traceable identifiers. This staged design allows for scalable dataset expansion and systematic evaluation across diverse driving environments.

The entire workflow is formalized in Algorithm~\ref{alg:AgentDrive_mcq}. Starting from a validated scenario JSON, the first stage generates a natural language description, constrained to 10–12 sentences, and enriched with contextual details, including road layout, environmental conditions, traffic controls, and dynamic vehicle maneuvers. The second stage applies an LLM-based synthesis process to convert the description into five structured MCQ JSONs, one per style. Each MCQ contains a concise question, four candidate answers, a single correct choice, and a textual rationale. Multiple validation checks (e.g., word length, option distinctness, label format, numeric presence when required, and action-orientation for comparative items) ensure structural consistency, with automated retries in place when failures occur. The final stage assembles the outputs into persistent MCQ objects, uniquely identified by content hashes, and stores them in the benchmark repository. This end-to-end design ensures traceability, reproducibility, and robustness of the generated dataset.

\subsection{Framework Illustration}
The overall architecture of the pipeline is depicted in Figure~\ref{fig:AgentDrive_mcq}. The process begins with the \texttt{AgentDrive} dataset, which encodes scenario attributes such as road layout, weather conditions, traffic density, and event triggers. These features are formalized into JSON files that serve as the structured representation of driving scenarios. A dedicated prompt builder then reformulates the structured data into prompts for a pool of LLMs, including DeepSeek-V3.1, GPT-4o, kimi-k2, and GPT-5o. The LLM pool generates narrative scenario descriptions and subsequently produces five reasoning-intensive MCQs per scenario, corresponding to the physics, policy, hybrid, scenario, and comparative styles. The intermediate outputs are stored as labeled JSON files and passed through a cross-check validation stage to ensure consistency and correctness. Finally, the validated items are aggregated into the \texttt{AgentDrive-MCQ} dataset, providing a robust benchmark for evaluating context-sensitive reasoning under multi-factor constraints.

\subsection{Prompt Design and Control}
Two carefully engineered prompts underpin the pipeline. The first, the \emph{Description Prompt}, instructs the model to summarize the scenario in 10–12 sentences, covering road geometry, environmental conditions, traffic controls, and dynamic agent behaviors. The result is a compact, self-contained narrative that encodes all relevant constraints. The second, the \emph{MCQ Prompt}, requires the model to generate a reasoning question of high difficulty, with exactly four candidate answers and one correct choice. Depending on the style, additional constraints enforce numeric presence (physics/hybrid), policy interpretation (policy), qualitative reasoning (scenario), or action comparison (comparative). Strict JSON formatting is imposed to enable automatic parsing and reproducibility. This design ensures that question difficulty emerges organically from scenario complexity and style-specific constraints, rather than from arbitrary distractor construction.

The pipeline yields two structured outputs. The first is a description-augmented scenario JSON that retains the original structured fields while enriching them with natural-language descriptions. The second is a set of \texttt{AgentDrive-MCQ} JSONs, each corresponding to one of the five reasoning styles. Each JSON includes a description, a reasoning-intensive question, four candidate answers, the correct answer, and a supporting rationale. This dual representation enables joint use in simulation-based evaluation and reasoning-based benchmarking.

\begin{figure}[!htbp]
    \centering
    \includegraphics[width=0.5\textwidth]{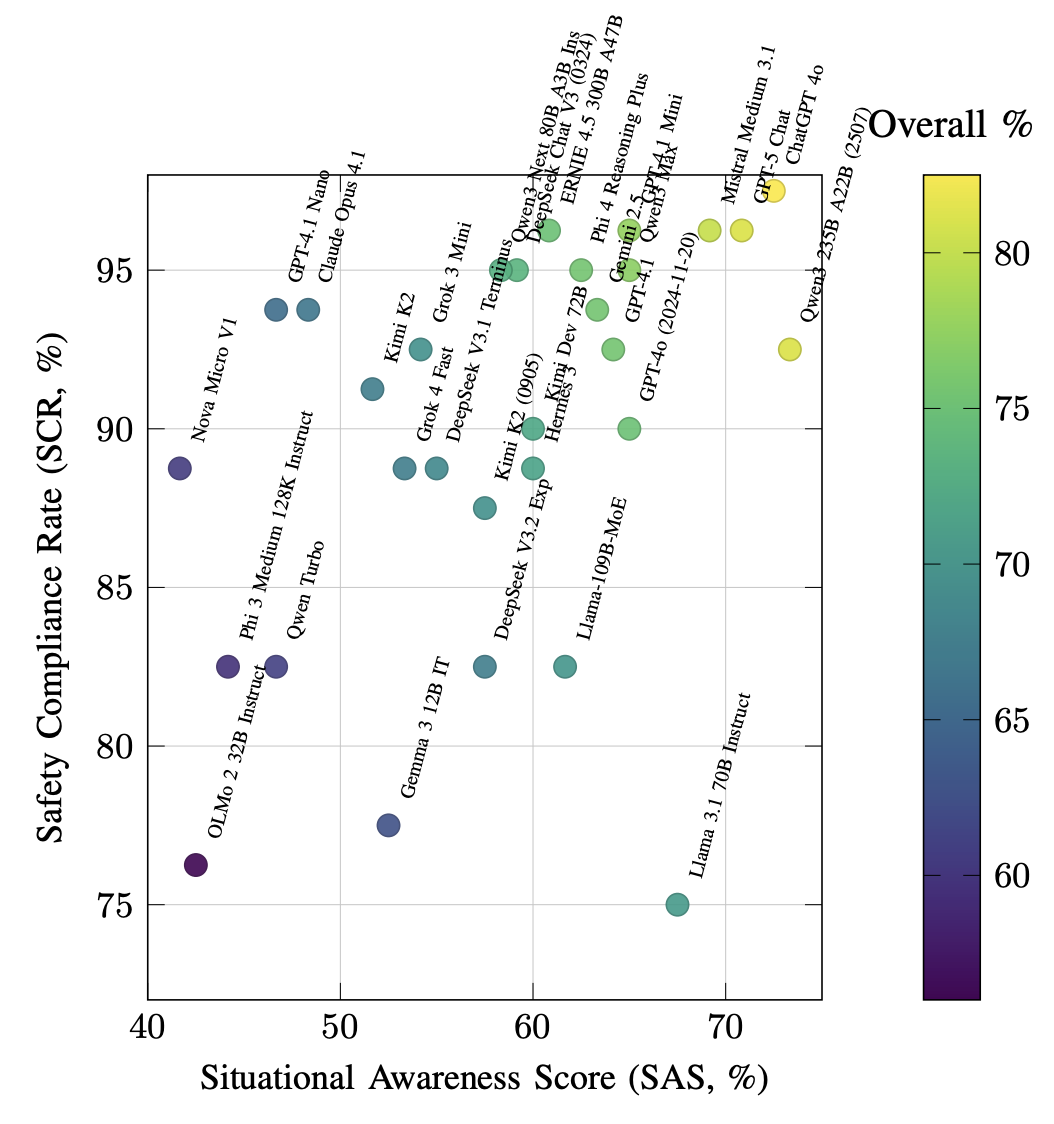}
    \caption{Top models by highest Overall: SAS vs. SCR.}
    \label{fig:SAS}
\end{figure}

\definecolor{cyanblue}{RGB}{224,238,255}

\begin{table*}[t]
\centering
\caption{Accuracy (\%) results of 50 examined LLM reasoning models evaluated across multiple reasoning styles using 2k samples from \texttt{AgentDrive-MCQ}.}
\renewcommand{\arraystretch}{0.7}
\rowcolors{2}{white}{cyanblue!70} 
\label{tab:llm-benchmark}
\scriptsize
\begin{tabular}{
l l l l
S[table-format=3.1, table-text-alignment=center]
S[table-format=3.1, table-text-alignment=center]
S[table-format=3.1, table-text-alignment=center]
S[table-format=3.1, table-text-alignment=center]
S[table-format=3.1, table-text-alignment=center]
S[table-format=3.1, table-text-alignment=center]
}
\toprule
\rowcolor{gray!35}\textbf{Model} & \textbf{Company} & \textbf{Size} & \textbf{License} &
\textbf{Comparative} & \textbf{Hybrid} & \textbf{Physics} & \textbf{Policy} & \textbf{Scenario} & \textbf{Overall} \\
\midrule
ChatGPT 4o & OpenAI & N/A & Proprietary & 90.0 & \maxnum{72.5} & 55.0 & \maxnum{100} & 95.0 & \maxnum{82.5} \\ \hline
GPT-5 Chat & OpenAI & N/A & Proprietary & 92.5 & 70.0 & 50.0 & \maxnum{100} & 92.5 & 81.0 \\ \hline
Qwen3 235B A22B (2507) & Alibaba & 235B & Open & 92.5 & 60.0 & \maxnum{67.5} & 87.5 & \maxnum{97.5} & 81.0 \\ \hline
Mistral Medium 3.1 & Mistral AI & N/A & Proprietary & \maxnum{95.0} & 60.0 & 52.5 & 97.5 & 95.0 & 80.0 \\ \hline
GPT-4.1 Mini & OpenAI & N/A & Proprietary & 90.0 & 55.0 & 50.0 & 95.0 & \maxnum{97.5} & 77.5 \\ \hline
GPT-4.1 & OpenAI & N/A & Proprietary & 90.0 & 50.0 & 52.5 & 90.0 & 95.0 & 75.5 \\ \hline
Phi 4 Reasoning Plus & Microsoft & 14B & Open & \maxnum{95.0} & 45.0 & 47.5 & 92.5 & \maxnum{97.5} & 75.5 \\ \hline
Gemini 2.5 Flash & Google & 391B & Proprietary & 92.5 & 65.0 & 32.5 & 92.5 & 95.0 & 75.5 \\ \hline
Qwen3 Max & Alibaba & N/A & Open & 95.0 & 52.5 & 47.5 & 95.0 & 95.0 & 77.0 \\ \hline
ERNIE 4.5 300B A47B & Baidu & 300B & Open & 85.0 & 45.0 & 52.5 & 95.0 & \maxnum{97.5} & 75.0 \\ \hline
DeepSeek Chat V3 (0324) & DeepSeek & 685B & Open & 77.5 & 55.0 & 45.0 & 95.0 & 95.0 & 73.5 \\ \hline
GPT-4o (2024-11-20) & OpenAI & N/A & Proprietary & 95.0 & 55.0 & 45.0 & 85.0 & 95.0 & 75.0 \\ \hline
Qwen3 Next 80B A3B Ins & Alibaba & 80B & Open & 82.5 & 52.5 & 40.0 & 95.0 & 95.0 & 73.0 \\ \hline
Kimi Dev 72B & Moonshot AI & 72B & Open & 80.0 & 57.5 & 42.5 & 87.5 & 92.5 & 72.0 \\ \hline
Hermes 3 (Llama 3.1 405B) & Nous Research & 405B & Open & 82.5 & 40.0 & 57.5 & 87.5 & 90.0 & 71.5 \\ \hline
Llama 3.1 70B Instruct & Meta & 70B & Open & 92.5 & 52.5 & 57.5 & 52.5 & \maxnum{97.5} & 70.5 \\ \hline
Llama-109B-MoE & DeepCogito & 109B & Open & 87.5 & 52.5 & 45.0 & 67.5 & \maxnum{97.5} & 70.0 \\ \hline
Kimi K2 (0905) & Moonshot AI & 1T & Open & 85.0 & 50.0 & 37.5 & 80.0 & 95.0 & 69.5 \\ \hline
Grok 3 Mini & xAI & N/A & Proprietary & 82.5 & 42.5 & 37.5 & 87.5 & \maxnum{97.5} & 69.5 \\ \hline
DeepSeek V3.1 Terminus & DeepSeek & N/A & Open & 85.0 & 32.5 & 47.5 & 85.0 & 92.5 & 68.5 \\ \hline
Grok 4 Fast & xAI & N/A & Proprietary & 87.5 & 30.0 & 42.5 & 87.5 & 90.0 & 67.5 \\ \hline
Kimi K2 & Moonshot AI & 1T & Open & 85.0 & 32.5 & 37.5 & 90.0 & 92.5 & 67.5 \\ \hline
DeepSeek V3.2 Exp & DeepSeek & N/A & Open & 85.0 & 32.5 & 55.0 & 75.0 & 90.0 & 67.5 \\ \hline
Claude 3.7 Sonnet (Thinking) & Anthropic & N/A & Proprietary & 92.5 & 40.0 & 65.0 & 42.5 & 95.0 & 67.0 \\ \hline
Claude Opus 4.1 & Anthropic & N/A & Proprietary & 92.5 & 42.5 & 10.0 & 95.0 & 92.5 & 66.5 \\ \hline
GPT-4.1 Nano & OpenAI & N/A & Proprietary & 87.5 & 17.5 & 35.0 & 97.5 & 90.0 & 65.5 \\ \hline
Gemma 3 12B IT & Google & 12B & Open & 57.5 & 50.0 & 50.0 & 70.0 & 85.0 & 62.5 \\ \hline
Llama 3.3 70B Instruct & Meta & 70B & Open & 90.0 & 47.5 & 55.0 & 27.5 & 95.0 & 63.0 \\ \hline
Qwen Turbo & Alibaba & N/A & Open & 87.5 & 22.5 & 30.0 & 75.0 & 90.0 & 61.0 \\ \hline
Nova Micro V1 & Amazon & N/A & Proprietary & 75.0 & 15.0 & 35.0 & 90.0 & 87.5 & 60.5 \\ \hline
Phi 3 Medium 128K Instruct & Microsoft & 14B & Open & 82.5 & 20.0 & 30.0 & 75.0 & 90.0 & 59.5 \\ \hline
Gemma 3 27B IT & Google & 27B & Open & 92.5 & 27.5 & 35.0 & 35.0 & 97.5 & 57.5 \\ \hline
OLMo 2 32B Instruct & AllenAI & 32B & Open & 57.5 & 25.0 & 45.0 & 62.5 & 90.0 & 56.0 \\ \hline
GPT-4o Mini & OpenAI & N/A & Proprietary & 85.0 & 20.0 & 52.5 & 30.0 & 90.0 & 55.5 \\ \hline
GPT-4o Mini (2024-07-18) & OpenAI & N/A & Proprietary & 85.0 & 20.0 & 52.5 & 25.0 & 90.0 & 54.5 \\ \hline
WizardLM 2 8×22B & Microsoft & 176B & Open & 60.0 & 45.0 & 45.0 & 25.0 & 90.0 & 53.0 \\ \hline
Claude Sonnet 4.5 & Anthropic & 468B & Proprietary & 75.0 & 0.0 & 0.0 & 85.0 & 95.0 & 51.0 \\ \hline
Llama 4 Scout & Meta & 17B & Open & 85.0 & 22.5 & 20.0 & 25.0 & 92.5 & 49.0 \\ \hline
LFM 3B & Liquid AI & 3B & Open & 67.5 & 25.0 & 40.0 & 12.5 & 92.5 & 47.5 \\ \hline
AFM 4.5B & Arcee AI & 4.5B & Open & 62.5 & 27.5 & 25.0 & 32.5 & 90.0 & 47.5 \\ \hline
Llama 4 Maverick & Meta & 17B & Open & 77.5 & 2.5 & 5.0 & 55.0 & 95.0 & 47.0 \\ \hline
Command R7B (12-2024) & Cohere & 7B & Open & 80.0 & 7.5 & 27.5 & 20.0 & 95.0 & 46.0 \\ \hline
Llama 3.1 8B Instruct & Meta & 8B & Open & 62.5 & 20.0 & 27.5 & 10.0 & 90.0 & 42.0 \\ \hline
Qwen 2.5 7B Instruct & Alibaba & 7B & Open & 47.5 & 12.5 & 20.0 & 50.0 & 57.5 & 37.5 \\ \hline
Llama 3.2 3B Instruct & Meta & 3B & Open & 52.5 & 7.5 & 35.0 & 7.5 & 80.0 & 36.5 \\ \hline
Llama 3.2 1B Instruct & Meta & 1B & Open & 62.5 & 10.0 & 30.0 & 7.5 & 72.5 & 36.5 \\ \hline
Mistral Nemo & Mistral AI & 12B & Open & 65.0 & 5.0 & 12.5 & 10.0 & 90.0 & 36.5 \\ \hline
GPT-3.5 Turbo Instruct & OpenAI & N/A & Proprietary & 52.5 & 5.0 & 27.5 & 5.0 & 82.5 & 34.5 \\ \hline
GLM 4.6 & Z-AI & N/A & Open & 60.0 & 5.0 & 2.5 & 45.0 & 37.5 & 30.0 \\ \hline
GPT-OSS-120B & OpenAI & 120B & Open & 22.5 & 5.0 & 10.0 & 10.0 & 10.0 & 11.5 \\ \hline
GPT-OSS-20B & OpenAI & 20B & Open & 17.5 & 7.5 & 0.0 & 15.0 & 7.5 & 9.5 \\
\bottomrule
\end{tabular} \\
\vspace{1mm}
\noindent\textit{LLM parameters:} \texttt{temperature = 0.0}, which controls randomness (0 = deterministic output); \texttt{max\_tokens = 16}, which defines the maximum number of tokens generated; \texttt{top\_p = 1.0}, which is the nucleus sampling parameter (1.0 = all tokens considered); and \texttt{max\_retries = 5}, which specifies the maximum number of retry attempts in case of LLM failure.
\end{table*}


\section{LLM Reasoning Performance on \texttt{AgentDrive-MCQ}}
\label{sec:performance}

The \texttt{AgentDrive-MCQ} benchmark provides a rigorous evaluation framework for assessing the reasoning capabilities of large language models across diverse driving-related scenarios. This section presents a comprehensive analysis of 50 leading LLMs, encompassing both proprietary and open-source architectures, and evaluates them across five reasoning dimensions: comparative, hybrid, physics, policy, and scenario. The goal is to identify strengths and weaknesses in model reasoning under varied environmental, physical, and ethical constraints. The analysis highlights how model scale, architecture, and training strategy influence performance, revealing that while frontier proprietary systems such as ChatGPT 4o and GPT-5 Chat dominate contextual and policy-based reasoning, advanced open models like Qwen3 235B A22B and ERNIE 4.5 300B A47B are rapidly closing the performance gap in structured, physics-grounded, and hybrid reasoning tasks. 

\subsection{Performance Metrics Definition}

The reasoning performance of each large language model (LLM) in \texttt{AgentDrive-MCQ} is evaluated across five distinct reasoning dimensions: \emph{Comparative}, \emph{Hybrid}, \emph{Physics}, \emph{Policy}, and \emph{Scenario}. Each dimension represents a specific cognitive ability required for robust, safety-aligned reasoning in autonomous driving contexts.

\paragraph{Overall Accuracy}

The \textbf{accuracy for each reasoning style} measures the proportion of correctly answered questions in that specific reasoning category:

\begin{equation}
\text{Accuracy}_{j} = 
\frac{\text{Correct Responses}_{j}}{\text{Total Questions}_{j}} \times 100, \\ [3pt]
\quad
\end{equation}

where $j$ $\in$ $\{\text{Comparative, Hybrid, Physics, Policy, Scenario}\}$

The \textbf{Overall Accuracy} serves as a unified indicator of model performance across all reasoning styles and is computed as:
\begin{equation}
\text{Overall Accuracy} =
\frac{1}{5}\sum_{j=1}^{5}\text{Accuracy}_{j}
\end{equation}
A higher overall accuracy reflects stronger reasoning consistency and generalization across heterogeneous driving scenarios.

Beyond these primary accuracies, two derived reliability metrics provide deeper insight into model safety and awareness.

\paragraph{Safety Compliance Rate (SCR)}

The \textbf{Safety Compliance Rate (SCR)} assesses the model's adherence to normative driving policies and ethical safety principles embedded in regulatory or operational constraints. It integrates both procedural correctness (\emph{Policy}) and behavioral consistency under realistic traffic situations (\emph{Scenario}):

\begin{equation}
\text{SCR} = 
\frac{\omega_{p} \cdot \text{Policy} + \omega_{s} \cdot \text{Scenario}}{\omega_{p} + \omega_{s}},
\end{equation}

where $\omega_{p}$ and $\omega_{s}$ denote the relative importance weights of the \emph{Policy} and \emph{Scenario} dimensions (by default, $\omega_{p} = \omega_{s} = 1$).  
A higher SCR reflects a model’s capability to comply with explicit safety regulations (e.g., traffic rules, right-of-way laws) and implicit ethical norms (e.g., harm minimization, risk-aware behavior) under uncertain or dynamic conditions. This metric thus serves as an indicator of \emph{safety alignment} and \emph{regulatory robustness} in autonomous decision-making.

\paragraph{Situational Awareness Score (SAS)}

The \textbf{Situational Awareness Score (SAS)} quantifies a model’s capacity to perceive, interpret, and predict environmental dynamics—a core cognitive competence in safety-critical autonomous reasoning. It aggregates perception–reasoning alignment across three complementary reasoning styles:

\begin{equation}
\text{SAS} =
\frac{\omega_{c} \cdot \text{Comparative} +
      \omega_{h} \cdot \text{Hybrid} +
      \omega_{f} \cdot \text{Physics}}{\omega_{c} + \omega_{h} + \omega_{f}},
\end{equation}

where $\omega_{c}$, $\omega_{h}$, and $\omega_{f}$ correspond to the relative weights assigned to \emph{Comparative}, \emph{Hybrid}, and \emph{Physics} reasoning dimensions, respectively (default: $\omega_{c} = \omega_{h} = \omega_{f} = 1$).  
A higher SAS denotes enhanced situational comprehension—capturing spatial awareness, causal inference, and physical feasibility reasoning. Models achieving strong SAS values exhibit robust contextual understanding and predictive foresight, especially in edge-case or high-risk driving scenarios.

\pgfplotstableread[col sep=comma]{
Model,Comparative,Hybrid,Physics,Policy,Scenario,Overall
ChatGPT 4o,90,72.5,55,100,95,82.5
GPT-5 Chat,92.5,70,50,100,92.5,81
Qwen3 235B A22B (2507),92.5,60,67.5,87.5,97.5,81
Mistral Medium 3.1,95,60,52.5,97.5,95,80
GPT-4.1 Mini,90,55,50,95,97.5,77.5
GPT-4.1,90,50,52.5,90,95,75.5
Phi 4 Reasoning Plus,95,45,47.5,92.5,97.5,75.5
Gemini 2.5,92.5,65,32.5,92.5,95,75.5
Qwen3 Max,95,52.5,47.5,95,95,77
ERNIE 4.5 300B A47B,85,45,52.5,95,97.5,75
DeepSeek Chat V3 (0324),77.5,55,45,95,95,73.5
GPT-4o (2024-11-20),95,55,45,85,95,75
Qwen3 Next 80B A3B Ins,82.5,52.5,40,95,95,73
Kimi Dev 72B,80,57.5,42.5,87.5,92.5,72
Hermes 3,82.5,40,57.5,87.5,90,71.5
Llama 3.1 70B Instruct,92.5,52.5,57.5,52.5,97.5,70.5
Llama-109B-MoE,87.5,52.5,45,67.5,97.5,70
Kimi K2 (0905),85,50,37.5,80,95,69.5
Grok 3 Mini,82.5,42.5,37.5,87.5,97.5,69.5
DeepSeek V3.1 Terminus,85,32.5,47.5,85,92.5,68.5
Grok 4 Fast,87.5,30,42.5,87.5,90,67.5
Kimi K2,85,32.5,37.5,90,92.5,67.5
DeepSeek V3.2 Exp,85,32.5,55,75,90,67.5
Claude 3.7 Sonnet (Thinking),92.5,40,65,42.5,95,67
Claude Opus 4.1,92.5,42.5,10,95,92.5,66.5
GPT-4.1 Nano,87.5,17.5,35,97.5,90,65.5
Gemma 3 12B IT,57.5,50,50,70,85,62.5
Llama 3.3 70B Instruct,90,47.5,55,27.5,95,63
Qwen Turbo,87.5,22.5,30,75,90,61
Nova Micro V1,75,15,35,90,87.5,60.5
Phi 3 Medium 128K Instruct,82.5,20,30,75,90,59.5
Gemma 3 27B IT,92.5,27.5,35,35,97.5,57.5
OLMo 2 32B Instruct,57.5,25,45,62.5,90,56
GPT-4o Mini,85,20,52.5,30,90,55.5
GPT-4o Mini (2024-07-18),85,20,52.5,25,90,54.5
}\AgentDrive

\pgfplotstablecreatecol[create col/expr={(\thisrow{Policy}+\thisrow{Scenario})/2}]{SCR}\AgentDrive
\pgfplotstablecreatecol[create col/expr={(\thisrow{Comparative}+\thisrow{Hybrid}+\thisrow{Physics})/3}]{SAS}\AgentDrive
\pgfplotstablecreatecol[create col/expr={100-(\thisrow{Policy}+\thisrow{Scenario})/2}]{IRI}\AgentDrive

\pgfplotstablesort[sort key=Overall, sort cmp=float >]{\SortedOverall}{\AgentDrive}
\pgfplotstablesort[sort key=IRI, sort cmp=float <]{\SortedIRI}{\AgentDrive}




\subsection{Overview of Model Performance}

The comparative and categorical analysis of 50 evaluated LLMs on the \texttt{AgentDrive-MCQ} benchmark reveals significant diversity in reasoning performance across five distinct dimensions: comparative, hybrid, physics, policy, and scenario. As summarized in Table~\ref{tab:llm-benchmark}, proprietary frontier models such as ChatGPT 4o (82.5\%) and GPT-5 Chat (81.0\%) from OpenAI dominated the benchmark, achieving perfect or near-perfect accuracy in policy (100\%) and scenario (97.5\%) reasoning tasks. These results underscore their superior contextual reasoning, ethical prioritization, and adaptability to complex decision-making scenarios. Among open-source systems, Qwen3 235B A22B reached a competitive 81.0\% overall accuracy, leading in physics-driven reasoning (67.5\%), while ERNIE 4.5 300B A47B achieved 75.0\%, reflecting the growing maturity of Chinese foundation models. Other strong performers, including Mistral Medium 3.1 (80.0\%) and GPT-4.1 Mini (77.5\%), demonstrated balanced reasoning proficiency across multiple domains, highlighting the value of large-scale fine-tuning and domain adaptation.

In contrast, smaller, earlier-generation models exhibited significant performance degradation. Models such as Llama 3.1 8B Instruct (42.0\%) and Qwen 2.5 7B Instruct (37.5\%) underperformed in multi-modal reasoning tasks, particularly when physical or contextual inference was required. The hybrid reasoning category emerged as the most challenging, with even top-tier models—ChatGPT 4o (72.5\%), GPT-5 Chat (70.0\%), and Qwen3 235B A22B (60.0\%)—struggling to integrate symbolic, numerical, and contextual information effectively. Conversely, mid-scale open models such as Kimi Dev 72B (72.0\%) and DeepSeek V3.1 Terminus (68.5\%) displayed encouraging stability, demonstrating that efficient architectures can approach the performance of frontier proprietary models when supported by focused fine-tuning. Overall, the results confirm that model scale, training strategy, and architecture design are pivotal for achieving robust reasoning, with open ecosystems rapidly closing the gap in structured and physics-aware reasoning.

\subsection{SAS–SCR Correlation}

Fig. \ref{fig:SAS} visualizes the relationship between the Situational Awareness Score (SAS) and the Safety Compliance Rate (SCR) for the top models ranked by Overall Accuracy from Table~\ref{tab:llm-benchmark}. The upper-right region of the scatter plot denotes the ideal operational zone, where models achieve both high situational awareness and strong safety compliance.

The results clearly indicate that frontier proprietary models such as ChatGPT 4o, GPT-5 Chat, and Qwen3 235B A22B dominate overall performance, reaching accuracies above 80 \%. Their SCR values exceed 95 \%, and their SAS values remain in the 65–75 \% range, demonstrating balanced reasoning between environmental understanding and safe decision-making. These findings highlight the maturity of large-scale, instruction-tuned architectures that integrate reasoning and physics-based inference effectively.

Mid-tier open-source models, including Mistral Medium 3.1 and ERNIE 4.5 300B A47B, exhibit competitive safety compliance but somewhat lower SAS, suggesting stronger rule adherence but limited depth in contextual or physical reasoning. Such patterns emphasize that mastering safety alignment does not automatically yield situational adaptability. Overall, a positive correlation emerges between SCR and SAS—models that understand complex physical environments tend also to make safer decisions. The clustering of high-performing models in the upper-right quadrant of Fig. \ref{fig:SAS} validates that reasoning precision and safety compliance are complementary competencies.

\subsection{Physics-Style Challenges}

The most demanding physics-style questions tested the quantitative reasoning depth of the evaluated models, involving complex physical dependencies such as traction, slope gradients, and variable deceleration rates. Even the top-performing models—GPT-5 Chat (50.0\%), ChatGPT 4o (55.0\%), and Qwen3 235B A22B (67.5\%)—showed reduced accuracy, reflecting the inherent difficulty of reasoning about motion dynamics and braking trajectories under uncertainty. Despite their contextual fluency, these models often underestimated or overestimated stopping distances, revealing weaknesses in applying physical formulas to real-world conditions. The findings emphasize that LLMs, while linguistically proficient, still lack numerical grounding and consistency when dealing with multivariable safety equations critical to autonomous driving.

\subsection{Policy-Style Challenges}

The policy-style reasoning category primarily assessed the models' ability to interpret traffic rules, safety margins, and prioritization policies under constrained visibility or dynamic hazards. High-performing systems such as ChatGPT 4o and GPT-5 Chat achieved 100\% accuracy, demonstrating mastery in policy alignment and context-sensitive judgment. However, smaller open models like Llama 3.1 8B Instruct (10.0\%) and AFM 4.5B (32.5\%) frequently failed to generalize safety rules, often choosing aggressive or unsafe responses. These results underscore the persistent gap between linguistic pattern recognition and actionable policy reasoning, suggesting that lightweight models require targeted, safety-aware fine-tuning to achieve reliable decision-making performance in regulatory and ethical reasoning domains.

\subsection{Hybrid-Style Challenges}

The hybrid-style reasoning dimension emerged as the most universally difficult in all model families—proprietary, open, and mid-scale. It required the simultaneous processing of symbolic policy cues and quantitative calculations, especially in multi-constraint decision-making scenarios (e.g., braking on gravel, handling downhill deceleration, or reacting under adverse weather). Even top-tier models such as ChatGPT 4o (72.5\%) and GPT-5 Chat (70.0\%) exhibited variability, while open counterparts such as Qwen3 235B A22B (60.0\%), Kimi Dev 72B (57.5\%), and DeepSeek Chat V3 (55.0\%) struggled with precise multi-factor estimation. The observed pattern confirms that hybrid reasoning—requiring the fusion of conceptual policy understanding and numerical grounding—remains an unresolved challenge in current LLM architectures. Future model advancements must focus on improving cognitive compositionality and structured reasoning under uncertainty to achieve human-level reliability in hybrid decision environments.

\subsection{Comparative-Style Challenges}

The comparative-style challenges assessed each model’s capacity to contrast multiple decision alternatives and identify the safest course of action under uncertain conditions. This dimension yielded generally strong results, with several models surpassing 90\% accuracy. Mistral Medium~3.1 and Phi~4~Reasoning~Plus achieved the highest comparative reasoning scores (95.0\%), followed closely by Qwen3~Max and GPT-5~Chat (92.5\%). The dominance of large proprietary systems such as ChatGPT~4o (90.0\%) and Gemini~2.5~Flash (92.5\%) underscores the importance of fine-tuned instruction alignment for maintaining decision consistency. In contrast, small and mid-scale open models—including Qwen~2.5~7B~Instruct (47.5\%) and Llama~3.1~8B~Instruct (62.5\%)—struggled with comparative reasoning, often failing to weigh competing outcomes accurately. These results suggest that comparative reasoning, while linguistically less demanding than numerical estimation, still benefits from high-capacity contextual modeling and refined safety-aware alignment. Overall, this dimension highlights the maturity of top-tier LLMs in structured choice-making but also reveals that smaller open models lack the inferential depth required to handle multi-option trade-offs reliably.

\subsection{Scenario-Style Challenges}

Scenario-style challenges required a holistic understanding of complex, dynamic environments with multiple interacting factors such as weather, visibility, and traffic density. This category produced some of the highest individual accuracies across the benchmark. Models like ChatGPT~4o, GPT-5~Chat, Phi~4~Reasoning~Plus, Qwen3~Max, and ERNIE~4.5~300B~A47B each achieved near-perfect scenario reasoning scores (97.5\%), demonstrating exceptional situational awareness and context synthesis. These systems consistently identified safe maneuvers by integrating spatial, temporal, and regulatory cues. In contrast, mid-tier models such as DeepSeek~Chat~V3 (95.0\%) and Kimi~Dev~72B (92.5\%) remained strong but slightly less consistent in handling unexpected hazards or environmental uncertainty. Smaller models—such as Llama~3.2~3B~Instruct (80.0\%) or Gemma~3~12B~IT (85.0\%)—showed substantial performance degradation, often missing implicit contextual dependencies. The overall trend indicates that scenario-based reasoning is the most stable and generalizable dimension for frontier models, as it combines perception, rule-following, and reasoning depth. However, achieving consistent real-world interpretability under diverse environmental conditions remains an open research challenge for the broader LLM ecosystem.

\section{Conclusion}
\label{sec:conclusion}

In this paper, we presented \texttt{AgentDrive}, a comprehensive open benchmark dataset designed to advance the training, fine-tuning, and evaluation of agentic AI systems in autonomous driving. Built upon a large-scale corpus of 300,000 LLM-generated driving scenarios, \texttt{AgentDrive} introduced a factorized scenario space that captured the complexity and diversity of real-world environments through seven orthogonal dimensions: scenario type, driver behavior, environment, road layout, objective, difficulty, and traffic density. An LLM-driven prompt-to-JSON specification pipeline ensured semantic richness, physical validity, and simulation readiness, while surrogate safety metrics and rule-based labeling provided interpretable ground truth for supervised learning and reasoning analysis.

To complement simulation-based evaluation, we proposed \texttt{AgentDrive-MCQ}, a reasoning-oriented benchmark containing 100,000 multiple-choice questions designed to assess the cognitive, ethical, and physics-grounded reasoning capabilities of LLM-based agents. We conducted a large-scale evaluation of fifty state-of-the-art LLMs—including GPT-5, ChatGPT 4o, Gemini 2.5 Flash, DeepSeek V3, Qwen3 235B, ERNIE 4.5 300B, Grok 4, and Mistral Medium 3.1—which demonstrated that while proprietary models currently dominate in contextual and policy reasoning, advanced open models rapidly closed the gap in structured and physics-based reasoning.

\texttt{AgentDrive} and \texttt{AgentDrive-MCQ} together provided a framework for benchmarking and improving agentic AI reasoning in safety-critical autonomous systems. Beyond evaluation, these resources enabled large-scale model training, fine-tuning, and transfer learning across simulation and reasoning domains. In future work, we plan to extend \texttt{AgentDrive} to multi-agent and multimodal environments, integrate real-world sensor data, and explore alignment strategies to further enhance the reliability and interpretability of LLM-driven autonomous agents.

To support open science and reproducibility, we released the complete \texttt{AgentDrive} dataset, labeled scenarios, \texttt{AgentDrive-MCQ} benchmark, Google Colab evaluation scripts, and all related materials on GitHub.

\bibliographystyle{IEEEtran}
\bibliography{bibliography} 

@inproceedings{ma2024lampilot,
  title={Lampilot: An open benchmark dataset for autonomous driving with language model programs},
  author={Ma, Yunsheng and Cui, Can and Cao, Xu and Ye, Wenqian and Liu, Peiran and Lu, Juanwu and Abdelraouf, Amr and Gupta, Rohit and Han, Kyungtae and Bera, Aniket and others},
  booktitle={Proceedings of the IEEE/CVF conference on computer vision and pattern recognition},
  pages={15141--15151},
  year={2024}
}

@article{gao2025foundation,
  title={Foundation Models in Autonomous Driving: A Survey on Scenario Generation and Scenario Analysis},
  author={Gao, Yuan and Piccinini, Mattia and Zhang, Yuchen and Wang, Dingrui and Moller, Korbinian and Brusnicki, Roberto and Zarrouki, Baha and Gambi, Alessio and Totz, Jan Frederik and Storms, Kai and others},
  journal={arXiv preprint arXiv:2506.11526},
  year={2025}
}

@article{fruhwirth2025stsbench,
  title={STSBench: A Spatio-temporal Scenario Benchmark for Multi-modal Large Language Models in Autonomous Driving},
  author={Fruhwirth-Reisinger, Christian and Mali{\'c}, Du{\v{s}}an and Lin, Wei and Schinagl, David and Schulter, Samuel and Possegger, Horst},
  journal={arXiv preprint arXiv:2506.06218},
  year={2025}
}

@article{yao2025agents,
  title={AGENTS-LLM: Augmentative GENeration of Challenging Traffic Scenarios with an Agentic LLM Framework},
  author={Yao, Yu and Bhatnagar, Salil and Mazzola, Markus and Belagiannis, Vasileios and Gilitschenski, Igor and Palmieri, Luigi and Razniewski, Simon and Hallgarten, Marcel},
  journal={arXiv preprint arXiv:2507.13729},
  year={2025}
}

@article{lebioda2025requirements,
  title={Are Requirements Really All You Need? Using LLMs to Generate Configuration Code: A Case Study in Automotive Simulations},
  author={Lebioda, Krzysztof and Petrovic, Nenad and Pan, Fengjunjie and Zolfaghari, Vahid and Schamschurko, Andr{\'e} and Knoll, Alois},
  journal={IEEE Access},
  year={2025},
  publisher={IEEE}
}

@article{xu2024drivegpt4,
  title={Drivegpt4: Interpretable end-to-end autonomous driving via large language model},
  author={Xu, Zhenhua and Zhang, Yujia and Xie, Enze and Zhao, Zhen and Guo, Yong and Wong, Kwan-Yee K and Li, Zhenguo and Zhao, Hengshuang},
  journal={IEEE Robotics and Automation Letters},
  year={2024},
  publisher={IEEE}
}

@inproceedings{chen2024driving,
  title={Driving with llms: Fusing object-level vector modality for explainable autonomous driving},
  author={Chen, Long and Sinavski, Oleg and H{\"u}nermann, Jan and Karnsund, Alice and Willmott, Andrew James and Birch, Danny and Maund, Daniel and Shotton, Jamie},
  booktitle={2024 IEEE International Conference on Robotics and Automation (ICRA)},
  pages={14093--14100},
  year={2024},
  organization={IEEE}
}

@article{jin2024large,
  title={Large language model as parking planning agent in the context of mixed period of autonomous vehicles and Human-Driven vehicles},
  author={Jin, Yuping and Ma, Jun},
  journal={Sustainable Cities and Society},
  volume={117},
  pages={105940},
  year={2024},
  publisher={Elsevier}
}

@article{wei2025ad,
  title={AD\^{} 2-Bench: A Hierarchical CoT Benchmark for MLLM in Autonomous Driving under Adverse Conditions},
  author={Wei, Zhaoyang and Qiang, Chenhui and Jiang, Bowen and Han, Xumeng and Yu, Xuehui and Han, Zhenjun},
  journal={arXiv preprint arXiv:2506.09557},
  year={2025}
}

@inproceedings{cui2024survey,
  title={A survey on multimodal large language models for autonomous driving},
  author={Cui, Can and Ma, Yunsheng and Cao, Xu and Ye, Wenqian and Zhou, Yang and Liang, Kaizhao and Chen, Jintai and Lu, Juanwu and Yang, Zichong and Liao, Kuei-Da and others},
  booktitle={Proceedings of the IEEE/CVF winter conference on applications of computer vision},
  pages={958--979},
  year={2024}
}

@article{sah2025advancing,
  title={Advancing Autonomous Vehicle Intelligence: Deep Learning and Multimodal LLM for Traffic Sign Recognition and Robust Lane Detection},
  author={Sah, Chandan Kumar and Shaw, Ankit Kumar and Lian, Xiaoli and Baig, Arsalan Shahid and Wen, Tuopu and Jiang, Kun and Yang, Mengmeng and Yang, Diange},
  journal={arXiv preprint arXiv:2503.06313},
  year={2025}
}

@article{bai20243d,
  title={Is a 3d-tokenized llm the key to reliable autonomous driving?},
  author={Bai, Yifan and Wu, Dongming and Liu, Yingfei and Jia, Fan and Mao, Weixin and Zhang, Ziheng and Zhao, Yucheng and Shen, Jianbing and Wei, Xing and Wang, Tiancai and others},
  journal={arXiv preprint arXiv:2405.18361},
  year={2024}
}

@inproceedings{kong2024superalignment,
  title={A superalignment framework in autonomous driving with large language models},
  author={Kong, Xiangrui and Braunl, Thomas and Fahmi, Marco and Wang, Yue},
  booktitle={2024 IEEE Intelligent Vehicles Symposium (IV)},
  pages={1715--1720},
  year={2024},
  organization={IEEE}
}

@article{cai2024driving,
  title={Driving with regulation: Interpretable decision-making for autonomous vehicles with retrieval-augmented reasoning via llm},
  author={Cai, Tianhui and Liu, Yifan and Zhou, Zewei and Ma, Haoxuan and Zhao, Seth Z and Wu, Zhiwen and Ma, Jiaqi},
  journal={arXiv preprint arXiv:2410.04759},
  year={2024}
}

@inproceedings{cao2024maplm,
  title={Maplm: A real-world large-scale vision-language benchmark for map and traffic scene understanding},
  author={Cao, Xu and Zhou, Tong and Ma, Yunsheng and Ye, Wenqian and Cui, Can and Tang, Kun and Cao, Zhipeng and Liang, Kaizhao and Wang, Ziran and Rehg, James M and others},
  booktitle={Proceedings of the IEEE/CVF conference on computer vision and pattern recognition},
  pages={21819--21830},
  year={2024}
}

@misc{highway-env,
  author = {Leurent, Edouard},
  title = {An Environment for Autonomous Driving Decision-Making},
  year = {2018},
  publisher = {GitHub},
  journal = {GitHub repository},
  howpublished = {\url{https://github.com/eleurent/highway-env}},
}

@article{xie2025vlms,
  title={Are vlms ready for autonomous driving? an empirical study from the reliability, data, and metric perspectives},
  author={Xie, Shaoyuan and Kong, Lingdong and Dong, Yuhao and Sima, Chonghao and Zhang, Wenwei and Chen, Qi Alfred and Liu, Ziwei and Pan, Liang},
  journal={arXiv preprint arXiv:2501.04003},
  year={2025}
}

@article{yang2023llm4drive,
  title={Llm4drive: A survey of large language models for autonomous driving},
  author={Yang, Zhenjie and Jia, Xiaosong and Li, Hongyang and Yan, Junchi},
  journal={arXiv preprint arXiv:2311.01043},
  year={2023}
}

@inproceedings{tang2024test,
  title={Test large language models on driving theory knowledge and skills for connected autonomous vehicles},
  author={Tang, Zuoyin and He, Jianhua and Pe, Dashuai and Liu, Kezhong and Gao, Tao and Zheng, Jiawei},
  booktitle={Proceedings of the 19th Workshop on Mobility in the Evolving Internet Architecture},
  pages={1--6},
  year={2024}
}

@article{pei2025methodology,
  title={Methodology and Benchmark for Automated Driving Theory Test of Large Language Models},
  author={Pei, Dashuai and Wu, Yiwen and He, Jianhua and Liu, Kezhong and Chen, Mozi and Xiao, Xuedou and Zhang, Shengkai and Zheng, Jiawei},
  journal={IEEE Transactions on Intelligent Transportation Systems},
  year={2025},
  publisher={IEEE}
}

@article{zhang2025bench2advlm,
  title={Bench2ADVLM: A Closed-Loop Benchmark for Vision-language Models in Autonomous Driving},
  author={Zhang, Tianyuan and Jin, Ting and Wang, Lu and Liu, Jiangfan and Liang, Siyuan and Zhang, Mingchuan and Liu, Aishan and Liu, Xianglong},
  journal={arXiv preprint arXiv:2508.02028},
  year={2025}
}

@inproceedings{huang2024drivlme,
  title={Drivlme: Enhancing llm-based autonomous driving agents with embodied and social experiences},
  author={Huang, Yidong and Sansom, Jacob and Ma, Ziqiao and Gervits, Felix and Chai, Joyce},
  booktitle={2024 IEEE/RSJ International Conference on Intelligent Robots and Systems (IROS)},
  pages={3153--3160},
  year={2024},
  organization={IEEE}
}

@article{cui2024large,
  title={Large language models for autonomous driving (llm4ad): Concept, benchmark, experiments, and challenges},
  author={Cui, Can and Ma, Yunsheng and Yang, Zichong and Zhou, Yupeng and Liu, Peiran and Lu, Juanwu and Li, Lingxi and Chen, Yaobin and Panchal, Jitesh H and Abdelraouf, Amr and others},
  journal={arXiv preprint arXiv:2410.15281},
  year={2024}
}

@techreport{zhou2025benchmarking,
  title={Benchmarking Large Language Models for Motorway Driving Scenario Understanding},
  author={Zhou, Ji and Zhao, Yongqi and Yang, Aixi and Eichberger, Arno},
  year={2025},
  institution={SAE Technical Paper}
}

@article{chiu2025v2v,
  title={V2v-llm: Vehicle-to-vehicle cooperative autonomous driving with multi-modal large language models},
  author={Chiu, Hsu-kuang and Hachiuma, Ryo and Wang, Chien-Yi and Smith, Stephen F and Wang, Yu-Chiang Frank and Chen, Min-Hung},
  journal={arXiv preprint arXiv:2502.09980},
  year={2025}
}

\end{document}